\documentclass[10pt,twocolumn,letterpaper]{article}

\usepackage{cvpr}      

\usepackage{graphicx}
\usepackage{amsmath}
\usepackage{amssymb}
\usepackage{booktabs}
\usepackage{multirow}
\usepackage{multicol}
\usepackage{makecell}

\newcommand\blfootnote[1]{%
\begingroup
\renewcommand\thefootnote{}\footnote{#1}%
\addtocounter{footnote}{-1}%
\endgroup
}
\usepackage[pagebackref,breaklinks,colorlinks]{hyperref}

\usepackage[capitalize]{cleveref}
\crefname{section}{Sec.}{Secs.}
\Crefname{section}{Section}{Sections}
\Crefname{table}{Table}{Tables}
\crefname{table}{Tab.}{Tabs.}

\def\cvprPaperID{****} 
\def\confName{CVPR}
\def\confYear{2023}

\begin{document}

\title{Revisiting Pre-training in Audio-Visual Learning}
\author{
\textbf{Ruoxuan Feng}$^1$\textbf{, Wenke Xia}$^1$\textbf{, Di Hu}$^{1,*}$
\smallskip 
\\
$^1$Gaoling School of Artificial Intelligence, Renmin University of China, Beijing\\
fengruox@gmail.com, \{xiawenke2022, dihu\}@ruc.edu.cn
}


\maketitle
\blfootnote{* Corresponding author.}

\begin{abstract}
Pre-training technique has gained tremendous success in enhancing model performance on various tasks, but found to perform worse than training from scratch in some uni-modal situations. This inspires us to think: are the pre-trained models always effective in the more complex multi-modal scenario, especially for the heterogeneous modalities such as audio and visual ones? We find that the answer is No. Specifically, we explore the effects of pre-trained models on two audio-visual learning scenarios: cross-modal initialization and multi-modal joint learning. When cross-modal initialization is applied, the phenomena of ``dead channel'' caused by abnormal Batchnorm parameters hinders the utilization of model capacity. Thus, we propose Adaptive Batchnorm Re-initialization (ABRi) to better exploit the capacity of pre-trained models for target tasks. In multi-modal joint learning, we find a strong pre-trained uni-modal encoder would bring negative effects on the encoder of another modality. To alleviate such problem, we introduce a two-stage Fusion Tuning strategy, taking better advantage of the pre-trained knowledge while making the uni-modal encoders cooperate with an adaptive masking method. The experiment results show that our methods could further exploit pre-trained models' potential and boost performance in audio-visual learning. The source code is available at \url{https://github.com/GeWu-Lab/Revisiting-Pre-training-in-Audio-Visual-Learning}

\end{abstract}

\begin{figure*}[ht]
  \centering
  \begin{subfigure}{0.49\linewidth}
    \includegraphics[width=8.1cm,height=6.2cm]{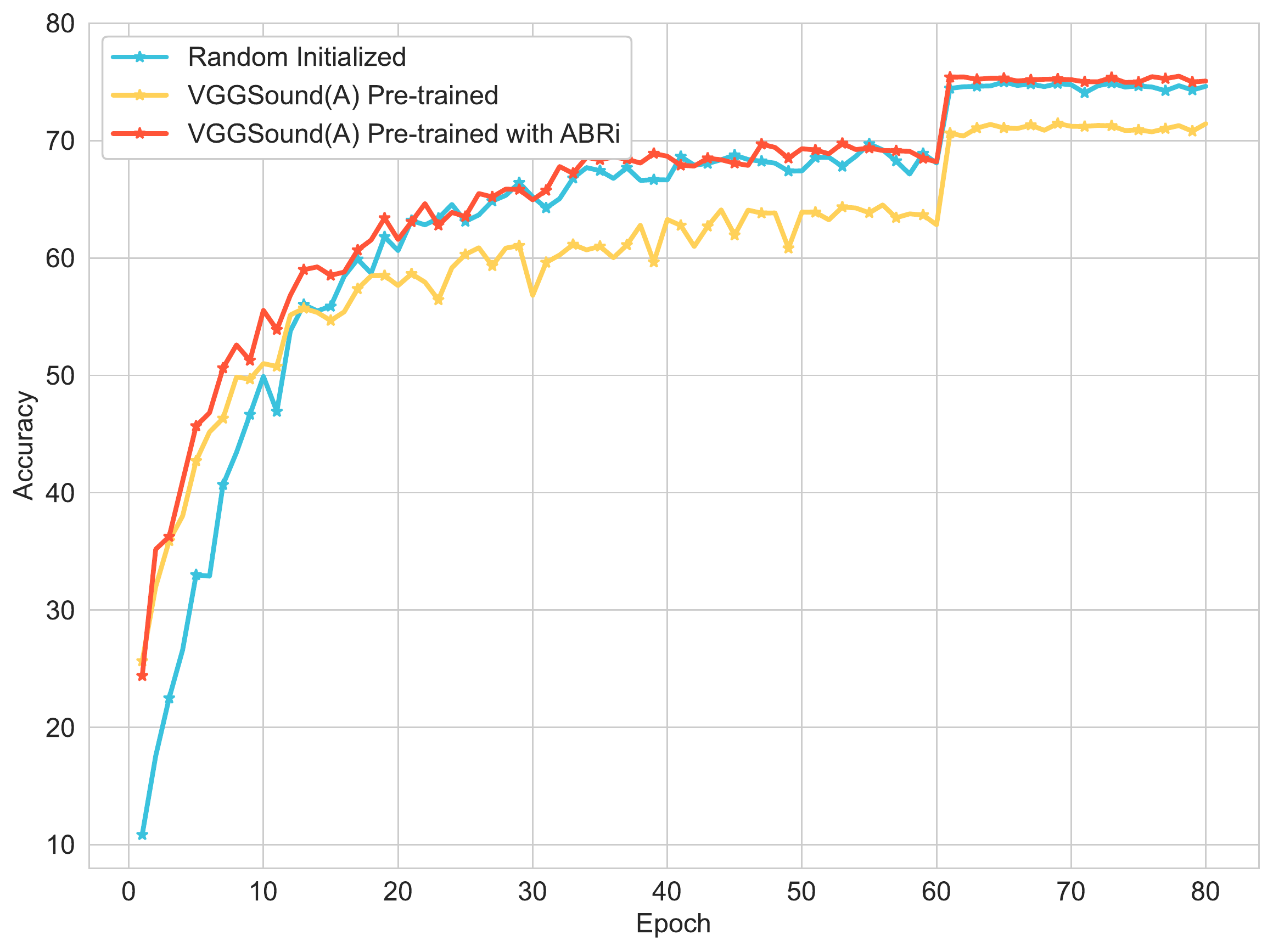}
    \caption{Accuracy of the models on CIFAR-100.}
    \label{fig:cifar-bn}
  \end{subfigure}
  \begin{subfigure}{0.49\linewidth}
  \includegraphics[width=8.40cm,height=6.2cm]{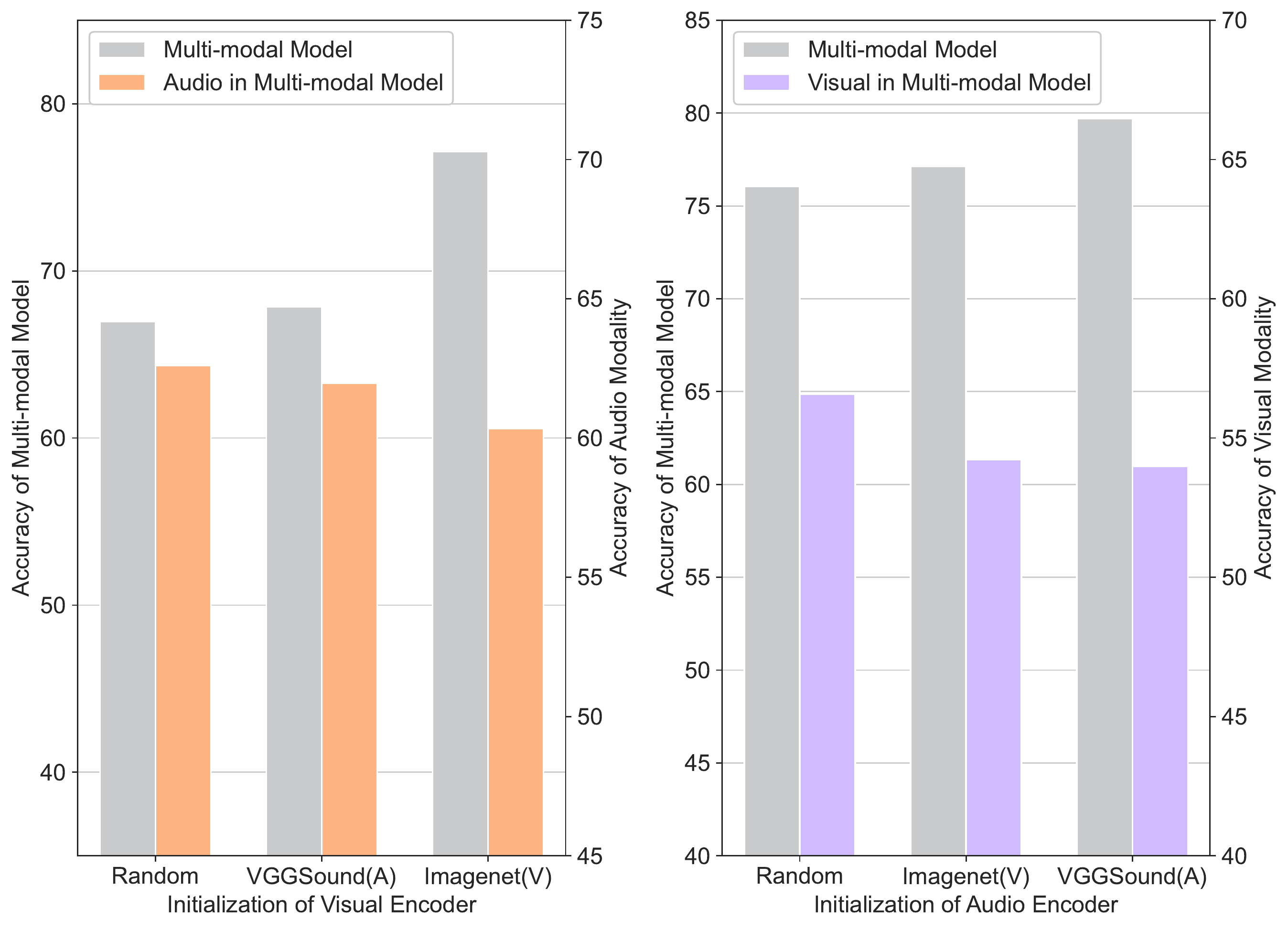}
    \caption{Accuracy of uni-modal and multi-modal models on Kinetics-Sounds.}
    \label{fig:linear probe}
  \end{subfigure}
  \caption{\textbf{Underutilization of pre-trained model in audio-visual learning}. (a) Performance of random initialized model, VGGSound~\cite{chen2020vggsound} pre-trained model and VGGSound pre-trained model with ABRi on CIFAR-100 testing set. (b) Performance of multi-modal models and linear-probing~\cite{alain2016understanding} results of uni-modal encoders in the multi-modal model. One modality uses different initialization while the other employs the same ImageNet pre-trained model on Kinetics-Sounds~\cite{arandjelovic2017look} testing set.}                            
  \label{fig:one}
\end{figure*}

\section{Introduction}

Pre-training technique promotes the development of deep learning and achieves promising results in many fields, such as computer vision~\cite{zhang2022dino,ghiasi2021simple}, acoustic processing~\cite{chen2022hts,zhang2020pushing} and natural language processing~\cite{radford2019language,raffel2020exploring}. The ability to extract features and encode representations learned on large-scale pre-training datasets is proved to be useful in the target task with a smaller scale of training samples~\cite{neyshabur2020whatisbeingtrans, yosinski2014transferable}.
However, when given stronger data augmentation~\cite{zoph2020rethinking}, more training samples~\cite{he2019rethinking,zoph2020rethinking} or different target tasks~\cite{he2019rethinking,zoph2020rethinking,li2019analysis}, researchers found that training from scratch occasionally outperforms training from pre-trained models, which implies pre-trained models could not always be profitable. 

In fact, the above findings are almost based on uni-modal scenarios. These lead to our way of thinking: for the much more complex multi-modal scenario, is the effectiveness of pre-trained models always held? To answer this, we choose to focus on a pair of typical heterogeneous modalities, audio and visual modality, where the heterogeneous data format is considered to bring more chances in exploring the effectiveness of pre-trained models. 
Concretely, we concentrate on two typical cases of audio-visual learning~\cite{wei2022avlearning}: \emph{cross-modal initialization} and \emph{multi-modal joint learning}.

\emph{Cross-modal initialization} is a technique used when a specific pre-trained model of the target modality could be hard to access. 
It employs the parameters of a pre-trained model of one modality as the initialization of the model for another modality~\cite{wang2016temporal}.
According to previous works, cross-modal initialization could improve model performance of the target modality, benefiting from the knowledge learned from another modality~\cite{palanisamy2020rethinking,kim2020urbanimagenet,amiriparian2020towardscnnaudio}.
These works empirically found that the filters of CNNs learned from one modality could be reused in target tasks of another modality to some extent, thus facilitating the learning of the model in target tasks. For example, the edge-aware filters in visual pre-trained model could help to detect the shape or edge in audio spectrograms~\cite{palanisamy2020rethinking}.
\emph{However, we find that the above reuse might not always bring performance gain in the target task, or is even worse than training from scratch, which is surprisingly contrary to common sense, as shown in \cref{fig:one}(a).}
Through multiple experiments, we point out that such phenomenon is brought by abnormal \emph{Batch Normalization} (Batchnorm)~\cite{ioffe2015batch} parameters in widely used Conv-BN-ReLU modules, which create ``dead channels\footnote{Dead channel indicates the channel that is almost inactivated regardless of the input sample.}'' and hinder the process of fine-tuning. 
To cope with this problem, we propose \emph{Adaptive Batchnorm Re-initialization} (ABRi) to offset the negative effect of abnormal Batchnorm parameters, which adaptively modifies each Batchnorm layer with additional initialized Batchnorm parameters.
The experiment results prove that ABRi accelerates the convergence speed and improves the performance when employing pre-trained model, as shown in \cref{fig:one}(a).

\emph{Multi-modal joint learning} is to train the encoders of different modality in the same framework simultaneously. In recent years, pre-training the uni-modal encoders of the multi-modal model on large-scale datasets (\textit{e}.\textit{g}., ImageNet~\cite{deng2009imagenet} and AudioSet~\cite{gemmeke2017audio}) then fine-tuning the model on multi-modal target tasks~\cite{van2022exploring,kamath2021mdetr,krishna2021using}, has become a dominant paradigm. Fine-tuning the pre-trained uni-modal encoders jointly in a multi-modal setting could achieve better performance~\cite{zeng2022pretrainboost,guo2021multipretrainboost}. However, little attention has been drawn to the possible negative impact between the encoders, which is brought by introducing the pre-trained uni-modal encoder for the multi-modal model. 
\emph{Concretely, we find that the representation quality of one modality in the multi-modal model drops, when a stronger pre-trained encoder is applied to the other modality, as shown in \cref{fig:one}(b). }
This interesting phenomenon indicates the uni-modal encoder could not fully exploit the knowledge obtained from the pre-training dataset. 
In order to perform effective cooperation among multiple uni-modal encoders while retaining their representation quality as much as possible, we propose a two-stage Fusion Tuning strategy. In the first stage, we individually fine-tune the uni-modal encoders to fully tap the potential of the pre-trained models. Then, in the second stage, we jointly fine-tune both encoders within the whole model, by adaptively performing sample-wise masking mechanism over each modality for better cooperation.
Using such Fusion Tuning strategy, we could achieve performance improvement on various audio-visual datasets.


We consider that the abnormal Batchnorm parameters and the drop in uni-modal representation quality are two aspects of the underutilization of pre-trained models, which could simultaneously hurt the performance of audio-visual learning. Hence, the proposed ABRi is combined with the Fusion Tuning strategy, which could jointly boost the performance of the multi-modal model further. More conducted experiments also shows that the proposed methods could promote more general learning scenario.
To summarize, we find that \textbf{the underutilization of not only model capacity, but also model knowledge limits the potential of the pre-trained model} in the multi-modal scenario. We hope these findings could inspire future works.

\section{Related works}

\begin{figure*}
    \begin{minipage}{0.37\linewidth}
        \centering
        \begin{subfigure}{1\linewidth}
            \includegraphics[width=6.37cm,height=5.6cm]{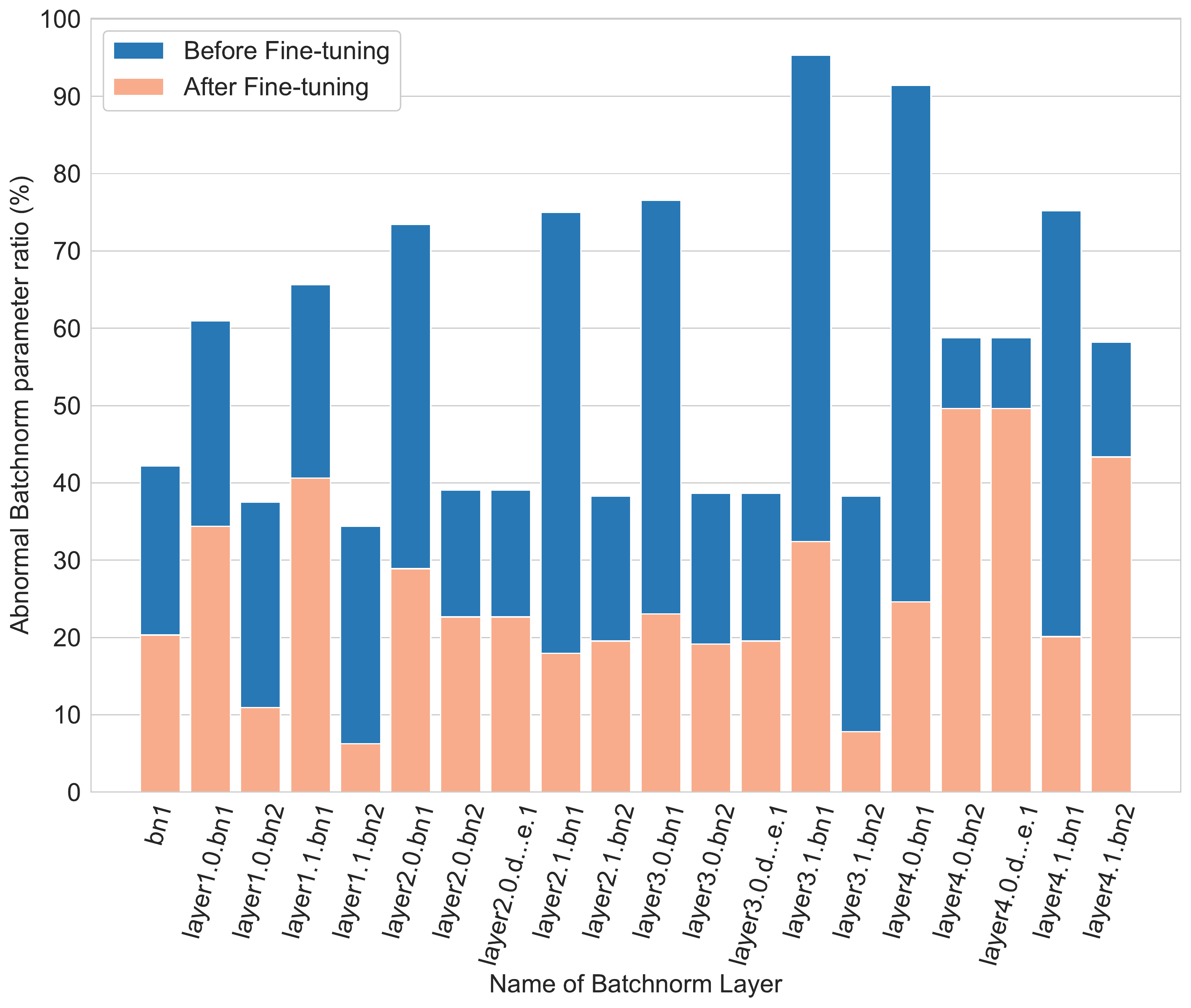}
            \caption{}
            \label{fig:dead-ratio}
        \end{subfigure}
    \end{minipage} 
    \medskip
    \begin{minipage}{0.28\linewidth}
        \centering
        \begin{subfigure}{1\linewidth}
            
            \includegraphics[width=4.8cm]{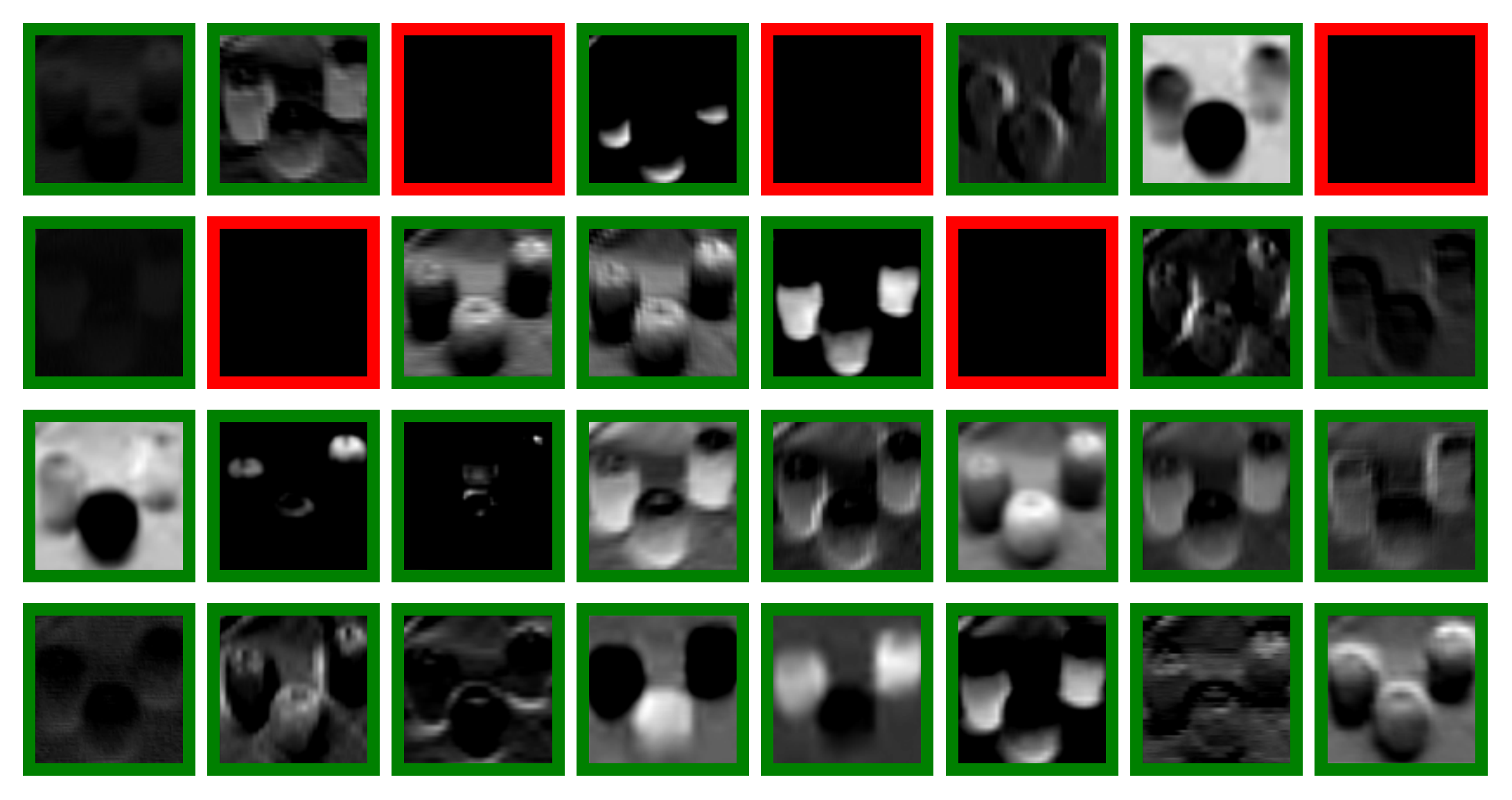}
            \caption{}
            \label{fig:dead-cifar}
        \end{subfigure}

        \begin{subfigure}{1\linewidth}
            
            \includegraphics[width=4.8cm]{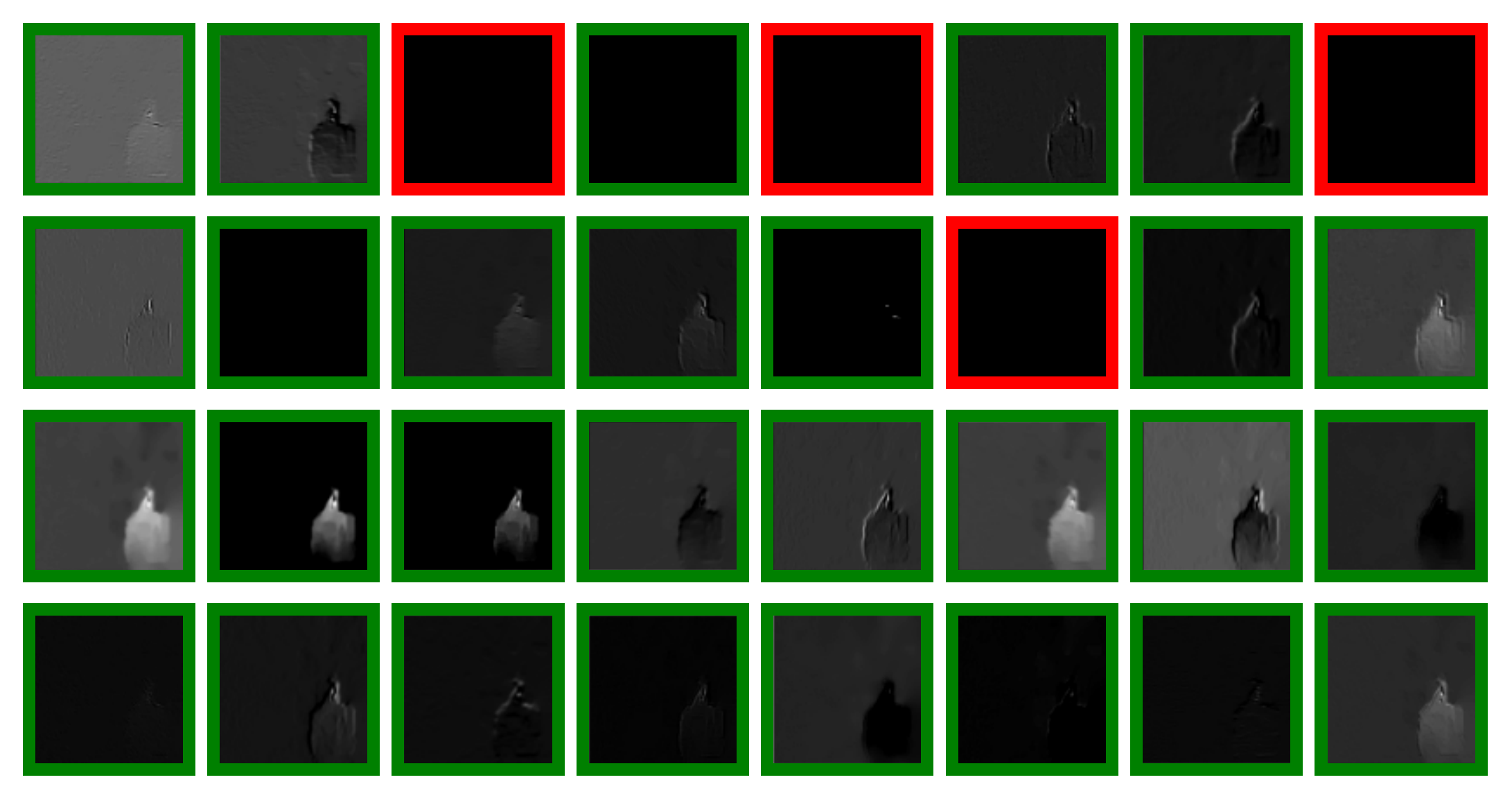}
            \caption{}
            \label{fig:dead-ucf}
        \end{subfigure}
    \end{minipage}
    \medskip
    \begin{minipage}{0.32\linewidth}
        \centering
        \begin{subfigure}{1\linewidth}
            
            \vspace{-0.02cm}
            \includegraphics[width=5.52cm]{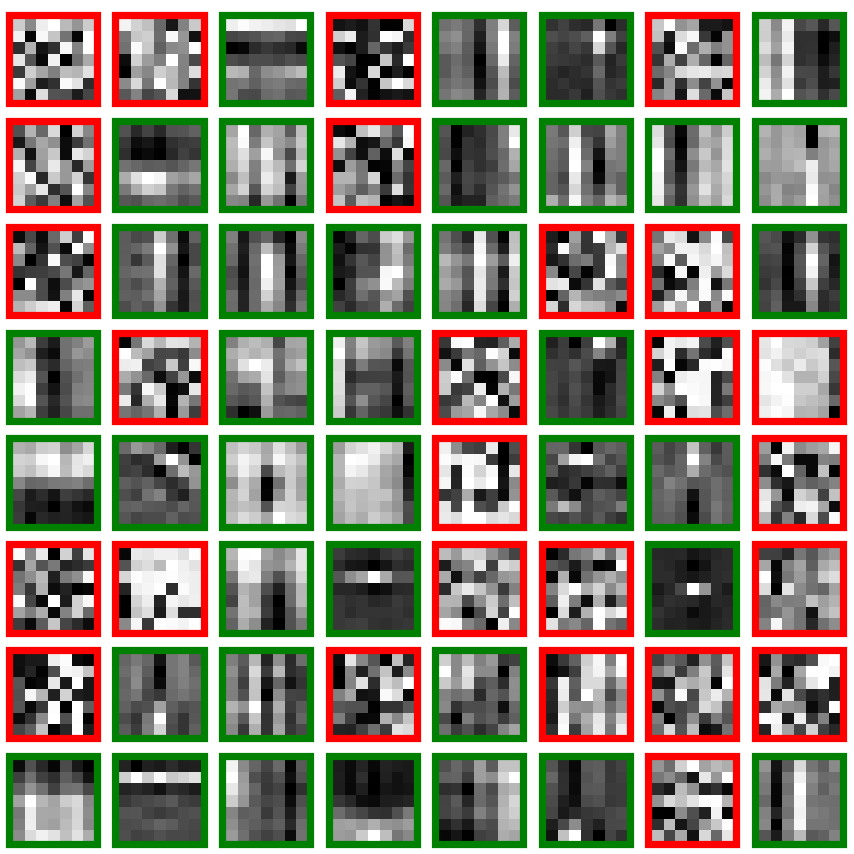}
            \vspace{0.02cm}
            \caption{}
            \label{fig:dead-filter}
        \end{subfigure}
    \end{minipage} 
    \vspace{-0.5cm}
\caption{\textbf{Visualization of dead channels phenomenon in  ResNet-18}. (a) Abnormal parameter ($<10^{-10}$) ratio of each Batchnorm layer in VGGSound pre-trained model before and after fine-tuned on CIFAR-100 dataset. (b) Feature maps after the first Conv-BN-ReLU in ImageNet pre-trained model. The feature maps of the channels marked with red boxes in the figure are dead features and the corresponding Batchnorm parameters are smaller than $2\times10^{-5}$.
This phenomenon is not unique to particular samples or classes, but exists for most samples. More examples are in \textit{Supp. Materials}. (c) Feature maps of optical flow after the first Conv-BN-ReLU in ImageNet pre-trained model fine-tuned on UCF-101\cite{soomro2012ucf101}. The channels with red boxes are remained dead channels.
(d) Filters of the first convolutional layer in VGGSound pre-trained model. The Batchnorm parameters corresponding to the filters marked with red boxes are smaller than $10^{-40}$.}
\label{fig:two}
\end{figure*}

\subsection{Pre-training and fine-tuning}
Researchers have come to recognize the effectiveness and potential of pre-trained models since the birth of the pre-training technique\cite{erhan2010does,devlin2018bert,hendrycks2019pretrainimpreove}.
Armed with the models pre-trained on large-scale datasets, recent works have achieved surprising results in computer vision~\cite{zhang2022dino,ghiasi2021simple}, acoustic processing~\cite{chen2022hts,zhang2020pushing} and natural language processing~\cite{radford2019language,raffel2020exploring}. The knowledge from pre-training datasets could be helpful even when faced with data in different domains~\cite{neyshabur2020whatisbeingtrans} or different tasks~\cite{girshick2014rich}.
However, recent works also found the limitation of pre-trained models. He \emph{et al.}~\cite{he2019rethinking} demonstrated that although ImageNet pre-training could accelerate convergence, it fails to improve the performance on a much different task. Zoph \emph{et al.}~\cite{zoph2020rethinking} found that pre-training may hurt performance when stronger augmentation is used during fine-tuning. These works analyzed the effectiveness of pre-training mainly in computer vision. In this paper, we focus on the potential limitation of pre-trained models fine-tuned on cross-modal and multi-modal tasks, which is rarely explored before.

\subsection{Cross-modal initialization}
Cross-modal initialization is a method that employs the pre-trained model of one modality as the initialization for the model of another modality. Previous works have proved that models pre-trained on large-scale datasets could provide a good initial point even if the modality of the target task is obviously different from the pre-training modality~\cite{gong2021ast,van2022exploring,palanisamy2020rethinking,huang2022flowformer,carreira2017quo}. 
When a specific pre-trained model of one modality is hard to access (\emph{e.g.}, audio and optical flow), it could be a potential choice to employ the easy-to-access pre-trained model of another modality (\emph{e.g.}, visual~\cite{simon2016imagenetzoo}). 
Accordingly, many recently proposed methods use ImageNet pre-trained model as the backbone for audio modeling and achieved considerable improvements~\cite{gong2021ast,kazakos2019epic,van2022exploring}. Van Horn \emph{et al.}~\cite{van2022exploring} also used ImageNet pre-trained models as the initialization of audio pre-training.
Attempting to explore why cross-modal initialization works even when the modalities are heterogeneous like audio and visual, preliminary research found that the well-trained edge-aware filters in the initial layers of ImageNet pre-trained models could be reused in audio related tasks~\cite{palanisamy2020rethinking}.
These works demonstrate that the knowledge of the pre-trained model could still be helpful in the target task of another modality. However, they did not investigate whether the capacity of pre-trained models is fully exploited, which is explored in our work.

\subsection{Multi-modal joint learning}
A typical training paradigm in various multi-modal (\emph{e.g.,} audio-visual) tasks is to use the information of both modalities simultaneously and fuse them to make prediction during the training phase~\cite{li2022learning,hu2021class}. Many multi-modal learning methods have been proposed to efficiently fuse the information in different levels, such as early fusion~\cite{castellano2008emotion,schuller2007audiovisual} and late fusion~\cite{alam2015late1,perera2011late2}. Hybrid fusion methods in multiple levels and with well-designed interactions further improve the performance of multi-modal joint learning~\cite{chen2016multi,nagrani2021attention}.

Aside from the fusion methods, some multi-modal joint learning mechanisms have been proposed to better train multi-modal networks. Wang \emph{et al.}~\cite{wang2020makes} found that the uni-modal over-fitting problem could degrade the performance of the joint model, then proposed Gradient-Blending across modalities to alleviate this problem. Peng \emph{et al.}~\cite{peng2022balanced} proposed an on-the-fly gradient modulation strategy to regulate the gradients of each uni-modal encoder dynamically. These works mainly focus on the cooperation between different modalities. However, they did not discuss the impact on cooperation, caused by applying well pre-trained encoders. In this work, we focus on exploring the influence between the uni-modal encoders when employing pre-training technique, then propose a two-stage Fusion Tuning strategy to reduce the impact.

\section{Cross-modal initialization}
\subsection{Underutilization of model capacity}
\label{sec:bn analysis}

\emph{Batch Normalization} (Batchnorm) has been widely adopted to improve the speed and stability of training deep neural networks~\cite{ioffe2015batch}. By introducing Batchnorm into \emph{convolutional neural network} (CNN) and combining it with \emph{convolutional layer} (Conv) and ReLU activation, the structure of Conv-BN-ReLU has become a general module in deep convolutional networks~\cite{he2016deep,jafari2020dru,chollet2017xception}. Specifically, the $k$-th channel of the Batchnorm layer essentially normalizes the $k$-th feature maps $x_k\in{\mathbb{R}^{M\times H\times W}}$ produced by the previous convolutional layer within $M$ mini-batch.
The normalized feature maps $\hat x_k$ of the $k$-th channel are then scaled via $\gamma_k$ and shifted via $\beta_k$ to obtain the final output $y_k$ as
\begin{equation}
\begin{aligned}
{y_k} = \gamma_k {\hat x_k} + \beta_k, 
\end{aligned}
\label{eq:bn}
\end{equation}
where $\gamma_k$ and $\beta_k$ are trainable parameters aiming to maintain the representation power of the network~\cite{ioffe2015batch}.  

However, we discover that the absolute value of some parameter pairs ($\lvert\gamma_k\rvert$, $\lvert\beta_k\rvert$) is significantly smaller than other parameters, after checking all the Batchnorm layers in VGGSound~\cite{chen2020vggsound} and ImageNet~\cite{deng2009imagenet} pre-trained  ResNet-18~\cite{he2016deep} models.
In \cref{fig:two}(a), blue bars indicate that these abnormal Batchnorm parameters could be found in every Batchnorm layer inside VGGSound pre-trained model. The abnormal ratio is even higher than $90\%$ in some layers. This unusual phenomenon attracts us to explore its impact on fine-tuning, especially when performing cross-modal initialization.

During feedforward, if $\lvert\gamma_k\rvert$ and $\lvert\beta_k\rvert$ are too small, the mean and variance of the corresponding feature maps are adjusted to tiny values due to \cref{eq:bn}. Worse still, in experiments we find that the channels with abnormal Batchnorm parameters are more likely to produce ``dead features\footnote{Dead features are feature maps with all zero entries.}''~\cite{zeiler2014visualizing} after ReLU. This phenomenon does not merely exist for particular samples or classes but for most samples, indicating the channels are hard to be activated, as shown in \cref{fig:two}(b). We name these channels as ``dead channels''.
The model performance could be limited if these dead channels remain after fine-tuning, as shown in \cref{fig:two}(c), for the corresponding feature maps provide less information to the next layer, causing information loss~\cite{mehta2018multilabel,liang2017informationloss}. 

During feedback, gradients back-propagated through Batchnorm layers are associated with the scaling parameter $\gamma_k$\footnote{$\beta_k$ has less impact in feedback phase.}:
\begin{footnotesize} 
\begin{equation}
\begin{aligned}
\frac{{\partial \mathcal{L}}}{{\partial {x_{k,m}}}} &= \frac{{\partial \mathcal{L}}}{{\partial {{\hat x}_{k,m}}}} \cdot \frac{1}{{\sqrt {\sigma _k^2} }} + \frac{{\partial \mathcal{L}}}{{\partial \sigma _k^2}} \cdot \frac{{2({x_{k,m}} - {\mu _k})}}{M} + \frac{{\partial \mathcal{L}}}{{\partial {\mu _k}}} \cdot \frac{1}{M}\\
&\propto \frac{{\partial \mathcal{L}}}{{\partial {{\hat x}_{k,m}}}}\\
&= \frac{{\partial \mathcal{L}}}{{\partial {y_{k,m}}}} \cdot \gamma_k ,
\end{aligned}
\label{eq:gradient}
\end{equation}
\end{footnotesize}
where $\mathcal{L}$ is the loss function and $m$ indicates the index of the
feature map within $M$ mini-batch. According to \cref{eq:gradient}, $\frac{{\partial \mathcal{L}}}{{\partial {x_{k,m}}}}$ is proportional to $\gamma_k$. This indicates that the abnormal $\gamma_k$ slows down the back-propagation of gradients. Thus, the corresponding filters in Conv-BN-ReLU structure are difficult to be updated. As shown in \cref{fig:two}(d), the filters marked with red boxes are obviously under-optimized.
\begin{figure}[t]
  \centering
  \includegraphics[width=8.3cm]{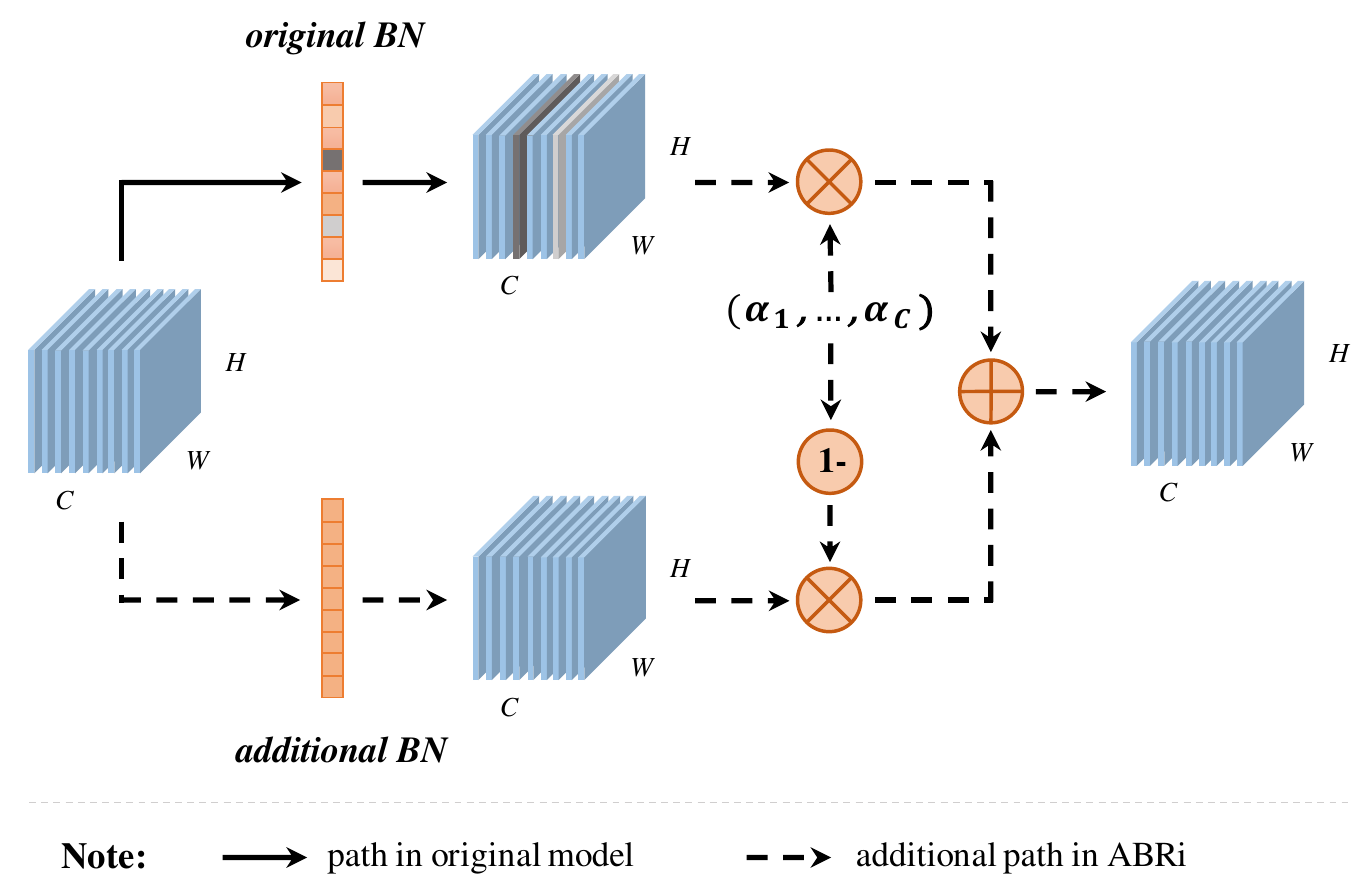}
  \caption{\textbf{The pipeline of Adaptive Batchnorm Re-initialization (ABRi).} The grey blocks in original Batchnorm layer indicate the channels with abnormal parameters.}
  \label{fig:pipeline of BN}
\end{figure}

Based on the above analysis, the abnormal Batchnorm parameters could bring potential problem during fine-tuning. Practically, the parameters of the pre-trained model require to adapt to the target task~\cite{neyshabur2020whatisbeingtrans,palanisamy2020rethinking,yosinski2014transferable}. Still, the abnormal parameters limit the update of some channels, resulting in the underutilization of the model capacity.
For the more challenging paradigm of cross-modal initialization, the model requires more tuning for a different modality.
Directly fine-tuning the pre-trained model in cross-modal scenario, we find the ratio of abnormal parameters could reduce to some extent, but still remains a high percentage, as the shown orange bars in \cref{fig:two}(a) and the red boxes in \cref{fig:two}(c). Hence, a kind of effective fine-tuning strategy is highly expected to reactivate all the ``dead channels'', then fully release the model capacity for the target task.

\begin{figure*}[ht]
  \centering
  \includegraphics[width=17.5cm]{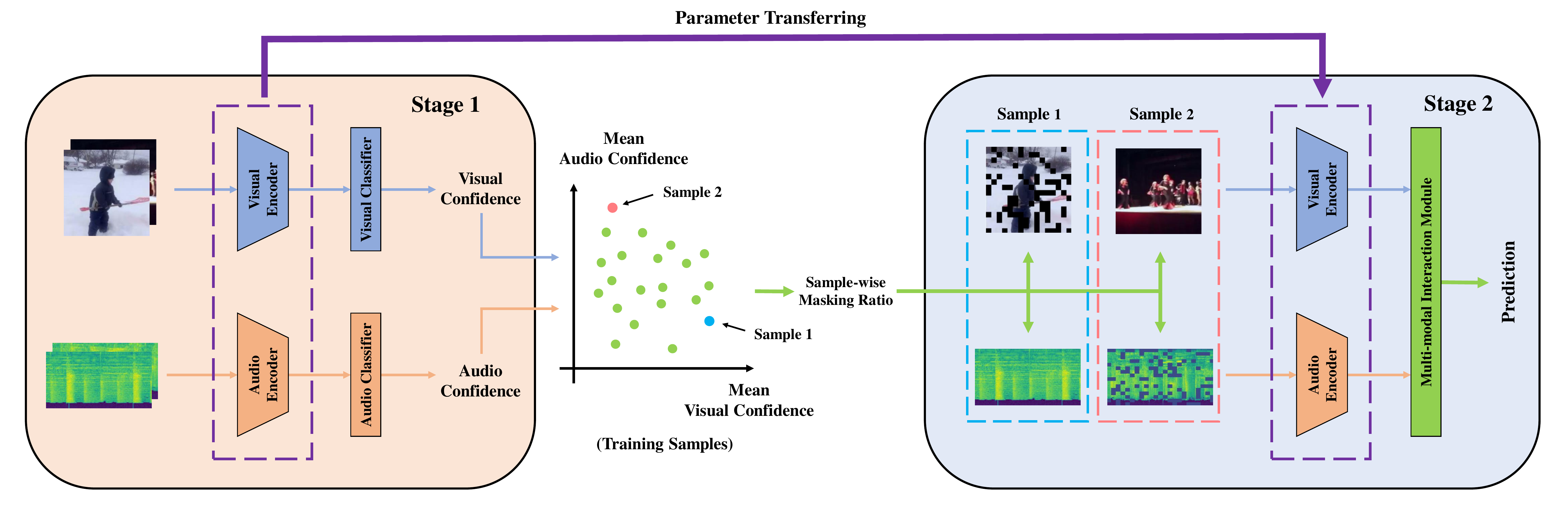}
  \caption{\textbf{The pipeline of two-stage Fusion Tuning strategy.} In stage 1, we separately train the uni-modal encoders and record the confidence of each training sample on both modalities. The sample-wise masking ratio is then calculated by the mean confidence of the sample on the two modalities. In stage 2, we train the entire multi-modal model consisting of the fine-tuned uni-modal encoders transferred from stage 1 and a multi-modal interaction module (\emph{e.g.,} a joint classifier). During the training process, we mask some parts of the easy-to-learn training sample of one modality based on the sample-wise masking ratio.}
  \label{fig:pipeline of FT}
\end{figure*}

\subsection{Adaptive Batchnorm Re-initialization}
\label{sec:abri}

To cope with the dead channel problem caused by the abnormal Batchnorm parameters, one naive strategy is to directly re-initialize every abnormal pair ($\gamma_k$ , $\beta_k$) to (1, 0) before training. However, this strategy could damage the target distribution of feature maps and affect the coordination with the next layer. To minimize the negative impact of abnormal parameters while ensuring coordination, an additional initialized Batchnorm layer is adaptively combined with each original Batchnorm layer. Specifically, the feature maps of the $k$-th channel go through the above two Batchnorm layers simultaneously:
\begin{equation}
\begin{aligned}
{y_k} = \alpha_k \cdot BN_{ori}(x_k) + (1 - \alpha_k) \cdot BN_{add}(x_k),
\end{aligned}
\end{equation}
where $BN_{ori}$ is the original Batchnorm layer, $BN_{add}$ is the additional Batchnorm layer and $\alpha_k$ is a trainable parameter to balance each original and additional channel. The pipeline is illustrated in \cref{fig:pipeline of BN}. The new Batchnorm layers help to reactivate the dead channels in the pre-trained model then the capacity of the model could be better utilized.

The total amount of the additional parameters is $1.5\times$ total BN parameters of the original model, which will not increase GPU memory occupation evidently. Furthermore, this method is not restricted to a particular modality and could be used in different modalities.

\section{Multi-modal joint learning}

\subsection{Underutilization of model knowledge}
\label{tag: joint training analysis}
Although introducing a stronger pre-trained encoder for one modality in the multi-modal model is very likely to improve the performance, we find that this could damage the representation ability of the other. As shown in \cref{fig:one}(b), both of the uni-modal encoders suffer a performance drop after introducing a stronger pre-trained model for another modality, indicating that both of the pre-trained encoders could have not yet adapted to the target task and thoroughly exploited the knowledge from the pre-trained model.

The above phenomenon indicates that the encoder of one modality could influence the learning of other encoder when performing joint training. Recent work preliminarily pointed out that high-quality predictions of one modality could reduce the gradient back-propagated to another encoder~\cite{peng2022balanced}, thus the optimization of the corresponding encoder could be insufficient and the representation quality could be reduced.
When directly initializing encoder with pre-trained model, predictions with high quality could be produced on the samples of one modality at the beginning, which are considered to be easy-to-learn. This could further exacerbate the above problem, making the encoders hard to 
effectively adapt to the target task and utilize task-oriented knowledge from pre-trained model.
Hence, how to coordinate the encoders while better utilizing the knowledge of the pre-training dataset for target task is a meaningful problem.

\subsection{Two-stage Fusion Tuning}

In order to address the above problem, we divide the process of joint training based on pre-trained models into two stages: separately training the uni-modal encoders then cooperatively fine-tuning the multi-modal network. The whole pipeline is shown in \cref{fig:pipeline of FT}.

In the first stage, to avoid the potential negative impact among modalities, we divide the multi-modal model into separated uni-modal ones, then fine-tune them on their own uni-modal dataset detached from the multi-modal dataset. 
This stage aims to make both of the encoders sufficiently exploit task-oriented knowledge for each modality.

In the second stage, the fine-tuned uni-modal encoders are assembled with a multi-modal interaction module to form the complete multi-modal model. The purpose of this stage is to perform effective cooperation among the separately trained encoders. However, the optimization of the multi-modal network, especially the interaction module, is still influenced by the easy-to-learn samples of one modality, reducing the cooperation quality.

\begin{table*}[ht]
\centering
\tabcolsep=0.34cm
\begin{tabular}{c|ccc|cc}
\toprule
\multirow{2}{*}{Pre-training dataset} &\multicolumn{3}{c|}{Audio tasks}&\multicolumn{2}{c}{Visual tasks}\\
 & ESC50 & TUT AS 2016 & Kinetics-Sounds & Cifar-100 & Caltech-256 \\
 \midrule[0.7pt]
Random & 77.95 & 62.78 & 54.56 & 75.31 & 48.09\\
ImageNet (V) & 89.35 & 72.10 & 62.07 & 79.97 & 77.55\\
VGGSound (A) & 90.35 & 73.89 & 63.53 & 72.54 & 47.00\\

\midrule
ImageNet (V)\dag &89.95 (+0.60)
 & 73.02 (+0.92)
 & 62.53 (+0.46)
 & \textbf{80.12 (+0.15)}
 & \textbf{77.86 (+0.31)} \\
VGGSound (A)\dag &\textbf{92.75 (+2.40)}
 & \textbf{75.37 (+1.48)}
 & \textbf{63.80 (+0.27)}
 & 75.58 (+3.06)
 & 49.69 (+2.69)\\

 \bottomrule
 
\end{tabular}
\caption{Comparison of performance with and without ABRi on different uni-modal datasets. \dag \ indicates ABRi is applied.}
\label{tab:res18 unimodal}
\end{table*}

Inspired by the data augmentation technique of adding noise~\cite{nazare2017deep,ahmadzadegan2021noise},
we propose to randomly mask some parts of the easy-to-learn samples. 
This strategy could lower the confidence on the easy-to-learn sample of one modality, reinforcing the learning of the hard-to-learn sample of the other modality and improving cooperating quality.
To obtain the learning difficulties of training samples on two modalities, we record the confidence of each training sample in every epoch of the first stage. We then calculate the mean confidence of each sample for each modality, which could reflect the modality-wise difficulty of learning to some extent.~\cite{swayamdipta2020dataset}.
The masking ratio $\{ m^a_i, m^v_i \}$ of the input $\{ x^a_i, x^v_i \}$ is then calculated by

\begin{footnotesize} 
\begin{equation}
\begin{aligned}
&m_i^a = \left\{ {\begin{array}{lr}
{\rho^a  \cdot \tanh [\eta^a \cdot (c_i^a - c_i^v)]} &c_i^a > c_i^v \ and\ c_i^v > t\\
0 &others,
\end{array}} \right. \\
&m_i^v = \left\{ {\begin{array}{lr}
{\rho^v  \cdot \tanh [\eta^v \cdot(c_i^v - c_i^a)]} &c_i^v > c_i^a \ and\ c_i^a > t\\
0 &others,
\end{array}} \right.
\end{aligned}
\end{equation}
\end{footnotesize}
where $\{ c^a_i, c^v_i \}$ is the mean confidence of the sample on two modalities, $\{ \rho^a, \rho^v \}$ and $\{ \eta^a, \eta^v \}$ are parameters to control the range of masking ratio and $t$ is a hyper-parameter to avoid masking the sample of one modality when the confidence on the other modality is apparently low (more likely to be noise). Our proposed ABRi technique in \cref{sec:abri} could be applied to the encoders in both stages, then the integrated method could further improve the performance of multi-modal joint learning.

\section{Experiments}
\subsection{Datasets}
\textbf{Pre-training datasets}. 
We select three widely used large-scale datasets: ImageNet-1K~\cite{deng2009imagenet} ($1.28M$ samples) VGGSound~\cite{chen2020vggsound} ($200K$ samples) and AudioSet~\cite{gemmeke2017audio} ($2.08M$ samples) to pre-train our models. ImageNet-1K is the pre-training dataset for visual modality, while VGGSound and AudioSet are pre-training datasets for audio modality.

\textbf{Uni-modal datasets}. For audio modality, we use ESC-50~\cite{piczak2015esc} and TUT Acoustic Scenes 2016~\cite{mesaros2016tut} datasets. For visual modality, CIFAR-100~\cite{krizhevsky2009learning} and Caltech-256~\cite{griffin2007caltech} are selected to be the target datasets. We also extract the audio modality from Kinetics-Sounds~\cite{arandjelovic2017look} as a uni-modal dataset.

\textbf{Multi-modal datasets}. For multi-modal tasks, Kinetics-Sounds, AVE~\cite{tian2018audio} and UCF-101~\cite{soomro2012ucf101} are chosen. We also randomly select a subset of VGGSound, where the amount of samples per class is 90,10,30 in training, validation and testing set. 
Totally, it contains 27,810 training samples, 3,090 validation samples and 9,270 testing samples. The amount of samples in each class is apparently lower than the number of classes, which could facilitate the evaluation on the utilization of pre-trained knowledge. 

\subsection{Experimental settings}
In our experiments, we employ ResNet-18 as the CNN backbones. For audio modality, we convert each 10-second audio waveform into 128-dimensional log Mel filterbank features using a window size of $25ms$ as~\cite{gong2021ast} does. Audio clips shorter than $10s$ are duplicated several times along the time axis and cut into $10s$. The obtained $128\times1024$ spectrogram is then duplicated three times to create a 3-channel input for the CNN network. In audio uni-modal tasks, we apply time mask and frequency mask. For the visual modality in multi-modal tasks, we extract frames of each video clip with 1 fps. Each video clip is averagely divided into 3 segments. Then, we randomly take 1 frames from each segment and put them into the 2D network as~\cite{zhao2018sound} does.

\subsection{Evaluation of ABRi on uni-modal tasks}

\begin{table*}[ht]
\centering
\tabcolsep=0.52cm
\begin{tabular}{c|c|ccc|ccc}
\toprule
\multirow{2}{*}{Dataset}&\multirow{2}{*}{Pre-train (Audio) }&\multicolumn{3}{c|}{Accuracy}&\multicolumn{3}{c}{mAP} \\
&&DF &JT  & FusT& DF & JT &FusT \\
 \midrule[0.7pt]
\multirow{2}{*}{VGGs}&AudioSet & 49.90 & 52.61 & \textbf{52.78}& 50.41 & 55.67&\textbf{55.89}\\
&ImageNet & 46.93  & 48.99& \textbf{49.62}& 47.33 & 50.77& \textbf{52.04}\\
 \midrule
\multirow{3}{*}{KS}&VGGSound & 76.01 & 79.13 & \textbf{81.29} & 81.98& 85.92& \textbf{87.23}\\
&AudioSet &79.13 & 81.02 & \textbf{83.90} & 84.82 & 88.03 &\textbf{90.20}\\
&ImageNet & 73.62 & 74.93 & \textbf{77.59} &79.88 & 81.23&\textbf{84.47}\\
 \bottomrule
 
\end{tabular}
\caption{Comparison with decision fusion and vanilla joint training on Kinetics-Sounds (KS) and VGGSound subset (VGGs). Visual encoders are pre-trained on ImageNet. DF means decision fusion, JT means joint training and FusT means our Fusion Tuning strategy.}
\label{tab:Fusion Tuning are good}
\end{table*}

\begin{table*}[t]
\centering
\tabcolsep=0.29cm
\begin{tabular}{c|c|ccc|cc|c}
\toprule
&Pre-train (Audio)&FusT* &FusT*-lr  & FusT-mean & OGM-GE\cite{peng2022balanced} & G-Blending\cite{wang2020makes} & FusT \\
 \midrule[0.75pt]
\multirow{3}{*}{Acc}&VGGSound &81.13&80.94&80.98&79.44&80.02&\textbf{81.29} \\
&AudioSet &83.52&83.37&83.13&80.79&83.29&\textbf{83.90}\\
&ImageNet & 74.82&75.20&75.93&75.66&76.93&\textbf{77.59}\\
 \midrule
\multirow{3}{*}{mAP}&VGGSound & 87.24&\textbf{87.38}&87.06&86.24&86.74&87.23\\
&AudioSet &90.15&89.91&89.82&88.41&90.08&\textbf{90.20}\\
&ImageNet &81.97&82.47&82.55&82.57&84.36&\textbf{84.47} \\
 \bottomrule
 
\end{tabular}
\caption{Comparison with variants of FusT and two modulation strategies on  Kinetics-Sounds. Visual encoders are pre-trained on ImageNet. FusT*, FusT*-lr and FusT-mean are three variants of FusT strategy. * indicates masking strategy is not applied in the second stage.}
\label{tab:baselines}
\end{table*}

To evaluate the effectiveness of ABRi, we conduct experiments on several uni-modal datasets. 
From the results in \cref{tab:res18 unimodal}, we could obtain four observations. 
Firstly, the pre-trained model of one modality is more effective in target tasks with the same modality. Concretely, the ImageNet pre-trained model outperforms the VGGSound pre-trained model on visual target tasks, while the result is the opposite on audio tasks.
Secondly, consistent performance improvement proves the effectiveness of ABRi. In addition, more notable improvements are visible in VGGSound pre-trained model, which has more abnormal Batchnorm parameters.
Thirdly, the improvement is more evident when performing cross-modal initialization. This phenomenon corresponds to the analysis in \cref{sec:bn analysis}. Specifically, the underutilization of model capacity brings more impact in the case of cross-modal initialization. Our proposed ABRi reactivates the dead channels and alleviates this problem.
Finally, it is noteworthy that the VGGSound pre-trained model with ABRi outperforms the model trained from scratch on visual datasets. Although it does not exceed the ImageNet pre-trained model having the same modality with the target task, this result demonstrates the knowledge from the audio modality could be helpful in the target task of visual modality with more informative features.

\subsection{Evaluation of Fusion Tuning strategy}
\label{sec:result on multi}

To evaluate the effect of our \emph{Fusion Tuning strategy} (FusT), we compare it with \emph{decision fusion} (DF) and \emph{vanilla joint training} (JT). In decision fusion, we average the softmax score of two modalities following~\cite{simonyan2014two}.
The result in \cref{tab:Fusion Tuning are good} shows a superior performance of our method among all pairs of pre-trained encoder on both datasets, demonstrating the effectiveness of our Fusion Tuning strategy. 

In order to show the advantage of the simple-wise masking strategy, we compare the Fusion Tuning strategy with three variants: (1) \emph{Fusion Tuning without masking} (FusT*), (2) \emph{Fusion Tuning without masking but with modality-specific learning rate}\footnote{We set the learning rate of audio encoders 0.5 times to visual encoders.} (FusT*-lr) and (3) \emph{Fusion Tuning with modality-wise masking ratio}\footnote{We compute the mean of the masking ratio for each modality and all the inputs of one modality share the same masking ratio.} (FusT-mean).
It is to be mentioned that the last two variants are modality-wise modulation methods, while ours is a more fine-grained sample-wise method.
As shown in \cref{tab:baselines}, our method outperforms all the variants on accuracy and most of the mAP. The accuracy of the models with AudioSet and VGGSound pre-trained audio encoder even dropped in FusT*-lr and FusT-mean variants, compared with Fusion Tuning without masking. This indicates that these easy but coarse variant strategies are not stable, demonstrating the meaning and advantage of our sample-wise modulation method.

We also make a comparison with two existing imbalance modulation strategies: OGM-GE~\cite{peng2022balanced} and Gradient-Blending~\cite{wang2020makes}.
OGM-GE is a batch-wise modulation strategy that dynamically regulates the gradient of uni-modal encoders based on the prediction within the mini-batch. Gradient-Blending is a modality-wise strategy that introduces additional uni-modal losses and adaptively modulates their weights.
As shown in \cref{tab:baselines}, our Fusion Tuning strategy shows superior performance among these methods. 
The above methods mainly focus on the cooperation between the two modalities, while less considering the utilization of model knowledge of each uni-modal encoder.
This indicates that the existing batch-wise and modality-wise methods could be less effective in handling the underutilization of model knowledge. In contrast, our sample-wise method exploits more potential of pre-trained models. 
\begin{table}[t]
\centering
\tabcolsep=0.39cm
\begin{tabular}{cccc}
\toprule
Strategy & Pre-train (Audio) & Acc & mAP \\
\midrule[0.7pt]
\multirow{3}{*}{FusT} & VGGSound & 81.29 & 87.23 \\
 & AudioSet & 83.90 & \textbf{90.20} \\
 & ImageNet & 77.59 & 84.47 \\
 \midrule
\multirow{3}{*}{FusT\dag}  & VGGSound & \textbf{81.48} & \textbf{87.34} \\
 & AudioSet & \textbf{84.14} & 89.99\\
 & ImageNet & \textbf{77.90} & \textbf{84.50}\\
 \midrule
 \multirow{3}{*}{FusT\ddag}  & VGGSound & 77.51  & 83.81 \\
 & AudioSet & 83.63  & 90.03 \\
 & ImageNet & 77.63  & 83.98 \\
 \bottomrule
 
\end{tabular}
\caption{Comparison of performance with and without ABRi on Kinetics-Sounds. Visual encoders are pre-trained on ImageNet. \dag \ indicates ABRi is applied on visual encoders. \ddag~indicates that ABRi is applied on both encoders.}
\label{tab:bn is good}
\end{table}

\subsection{Combination and extension of the two methods}
\label{sec:combine}
Based on the two aspects of underutilization of pre-trained models, model capacity and model knowledge, we combine ABRi with Fusion Tuning strategy and apply them on Kinetics-Sounds dataset.
Specifically, we conduct an experiment on Kinetics-Sounds dataset, comparing the performance of the multi-modal model without ABRi, the model with ABRi on visual encoder and the model with ABRi on both encoders. The slight boosts on the model with ABRi only on visual encoder shown in \cref{tab:bn is good} demonstrate that the dead channels hurt the performance to some extent, but the underutilization of model knowledge is the critical reason for the limited performance on Kinetics-Sounds. Surprisingly, the performance of the models drops when ABRi is also applied on audio encoders. 
In Kinetics-Sounds, audio samples are generally more easy-to-learn than visual samples. As a result, the optimization of the multi-modal model is dominated by audio modality, leading to under-optimization of visual modality~\cite{peng2022balanced,du2021improving}.
The experiment result indicates that simply improving the representation quality of the dominant modality (such as audio in this case) could aggravate the under-optimization problem of the other modality and hurt the performance in some situations. 
This severe problem could increase the difficulty of modulation, and is worthy of exploration in future works.

\textbf{Extending to more complex interaction modules.} We apply the two-stage Fusion Tuning strategy on two representative frameworks of audio-visual event localization task: Audio-Guided Visual Attention (AGVA)~\cite{tian2018audio} and Positive Sample Propagation (PSP)~\cite{zhou2021positive} for AVE dataset. 
Since audio-visual event localization is a more fine-grained task, these frameworks have well-designed cross-modal interaction mechanisms. Hence, we extract the parts only relevant to a single modality and train them in the first stage. In addition, we apply ABRi on both uni-modal encoders, for AVE dataset is a more balanced dataset. The complete model is then fine-tuned in the second stage based on the confidence recorded in the first stage.
The comparison results are shown in \cref{tab:AVE}. 
The improvement demonstrates that our Fusion Tuning strategy could be applicable with more complex interaction modules.

\textbf{Extending to other modalities.}
To demonstrate the generality of our methods on other modalities, we conduct an experiment on UCF-101 dataset. The two modalities are RGB and optical flow, which are less heterogeneous than audio and visual. 
The models for optical flow modality often use ImageNet pre-trained model as initialization, which could be influenced by the dead channels as well.
The performance gain shown in \cref{tab:UCF101} indicates the universality of our methods for different modalities.

\begin{table}[t]
\centering
\tabcolsep=0.26cm
\begin{tabular}{cccc}
\toprule
Framework&Strategy & Pre-train (Audio)&  Acc \\
\midrule[0.7pt]
\multirow{4}{*}{AGVA~\cite{tian2018audio}}&\multirow{2}{*}{JT} & VGGSound & 69.15  \\
& & ImageNet & 67.41  \\
\cmidrule(l){2-4}
 &\multirow{2}{*}{FusT\ddag} & VGGSound & \textbf{72.64}  \\
& & ImageNet & \textbf{68.41} \\
\midrule[0.6pt]
\multirow{4}{*}{PSP~\cite{zhou2021positive}}&\multirow{2}{*}{JT} & VGGSound & 73.38 \\
& & ImageNet  & 68.41  \\
\cmidrule(l){2-4}
 &\multirow{2}{*}{FusT\ddag} & VGGSound & \textbf{74.13} \\
& & ImageNet & \textbf{71.89} \\
 \bottomrule
 
\end{tabular}
\caption{Comparison on AVE dataset with and without Fusion Tuning strategy and ABRi methods in two frameworks. Visual encoders are pre-trained on ImageNet. \ddag \ indicates that ABRi is applied on both encoders. }
\label{tab:AVE}
\end{table}

\begin{table}[t]
\centering
\tabcolsep=0.42cm
\begin{tabular}{ccc}
\toprule
Strategy  & Accuracy& mAP \\
\midrule[0.7pt]
Decision Fusion & 81.49 & 86.49 \\
\midrule
Joint Training & 78.93 & 84.70 \\
\midrule
Fusion Tuning\ddag  & \textbf{83.62} & \textbf{89.00} \\
 \bottomrule
 
\end{tabular}
\caption{Performance of the model with two-stage Fusion Tuning strategy and ABRi on UCF-101. Encoders are pre-trained on ImageNet. \ddag \ indicates that ABRi is applied on both encoders.}
\label{tab:UCF101}
\end{table}

\section{Discussion}
In this paper, we find that the underutilization of model capacity and model knowledge limits the potential of pre-trained models in two typical cases of audio-visual learning and analyze their causes. We propose Adaptive Batchnorm Re-initialization (ABRi) and two-stage Fusion Tuning strategy to better utilizing the pre-trained models in both cases.

\textbf{Meanings of cross-modal initialization.} Although the VGGSound pre-trained model with ABRi outperforms training from scratch on visual tasks, it is left behind by the ImageNet pre-trained model. Our experiment is only an attempt and we do not recommend using the pre-trained model from another less informative modality if the pre-trained model of its modality exists. However, if the pre-trained model is hard to access, cross-modal initialization will become a good choice other than training from scratch. 

\textbf{Limitation.} Although our methods improve the performance of pre-trained models in uni-modal and multi-modal tasks, they require additional GPU memory and more training epochs. The purposed ABRi could contribute less for the model with few abnormal Batchnorm parameters. 

{\small
\bibliographystyle{ieee_fullname}
\bibliography{arxiv}
}

\clearpage
\appendix
\section{Analysis of dead channels}
\label{sec:intro}

\subsection{Gradient analysis}
\begin{table}[h]
\centering
\begin{tabular}{cc}
\toprule
Symbol & Meaning \\
\midrule[0.7pt]
$\mathcal{L}$ & Loss function\\
$M$ & Mini-batch size\\
$m$ & Index of the feature map within $M$ mini-batch\\
$x_{k,m}$&$m$-th feature map of the $k$-th channel\\
$\hat x_{k,m}$&$m$-th normalized feature maps of $k$-th channel\\
$\mu _k,\sigma _k^2$&Mean and variance of the $k$-th channel\\
$\gamma_k$ & Trainable scaling parameter in BN\\
\bottomrule
\end{tabular}
\caption{Symbols and their meanings.}
\label{tab:symbol}
\end{table}
During feedback, the gradient back-propagated through
Batchnorm\cite{ioffe2015batch} layers is
\begin{equation}
\begin{aligned}
\frac{{\partial \mathcal{L}}}{{\partial {x_{k,m}}}} &= \frac{{\partial \mathcal{L}}}{{\partial {{\hat x}_{k,m}}}} \cdot \frac{1}{{\sqrt {\sigma _k^2} }} + \frac{{\partial \mathcal{L}}}{{\partial \sigma _k^2}} \cdot \frac{{2({x_{k,m}} - {\mu _k})}}{M} \\
&+ \frac{{\partial \mathcal{L}}}{{\partial {\mu _k}}} \cdot \frac{1}{M},
\end{aligned}
\label{eq:gradient1}
\end{equation}
Every term in the gradient $\frac{{\partial \mathcal{L}}}{{\partial {x_{k,m}}}}$ is multiplied with $\frac{{\partial \mathcal{L}}}{{\partial {\hat x_{k,m}}}}$ for
\begin{equation}
\begin{aligned}
\frac{{\partial \mathcal{L}}}{{\partial \sigma _k^2}} &= \sum\limits_{m = 1}^M {\frac{{\partial \mathcal{L}}}{{\partial {{\hat x}_{k,m}}}} \cdot ({x_{k,m}} - {\mu _k})}  \cdot \frac{{ - 1}}{2}{(\sigma _k^2)^{ - \frac{3}{2}}}, \\
\frac{{\partial \mathcal{L}}}{{\partial {\mu _k}}} &= \sum\limits_{m = 1}^M {\frac{{\partial \mathcal{L}}}{{\partial {{\hat x}_{k,m}}}} \cdot \frac{{ - 1}}{{\sqrt {\sigma _k^2} }}} .
\end{aligned}
\label{eq:gradient2}
\end{equation}
However, $\frac{{\partial \mathcal{L}}}{{\partial {\hat x_{k,m}}}}$ is proportional to $\gamma_k$ via
\begin{equation}
\begin{aligned}
\frac{{\partial \mathcal{L}}}{{\partial {{\hat x}_{k,m}}}} &= \frac{{\partial \mathcal{L}}}{{\partial {y_{k,m}}}} \cdot \gamma_k .
\end{aligned}
\label{eq:gradient3}
\end{equation}
As a result, the gradient $\frac{{\partial \mathcal{L}}}{{\partial {x_{k,m}}}}$ is proportional to $\gamma_k$, indicating that the abnormal (very small in this ) $\gamma_k$ slows down the back-propagation of gradients. The meanings of the symbols above are in \cref{tab:symbol}.

\begin{table*}[ht]
\centering
\tabcolsep=0.42cm
\begin{tabular}{ccccc}
\toprule
Dataset&Strategy  & FLOPs per epoch& Epochs & Total FLOPs \\
\midrule[0.7pt]
\multirow{5}{*}{Kinetic-Sounds}& \multirow{2}{*}{Joint Training} & \multirow{2}{*}{$1.02\times10^{10}$} & \multirow{2}{*}{60} & \multirow{2}{*}{$6.15\times10^{11}$} \\
&&&&\\
\cmidrule{2-5}
&\multirow{3}{*}{Fusion Tuning}  & Stage 1-Audio: $4.77\times10^{9}$ & 40 &\multirow{3}{*}{$8.34\times10^{11}$}\\
&&Stage 1-Visual: $5.48\times10^{9}$ & 80\\
&&Stage 2: $1.02\times10^{10}$& 20 \\
\midrule
\multirow{5}{*}{AVE}& \multirow{2}{*}{Joint Training} & \multirow{2}{*}{$2.36\times10^{10}$} & \multirow{2}{*}{50} & \multirow{2}{*}{$1.18\times10^{12}$} \\
&&&&\\
\cmidrule{2-5}
&\multirow{3}{*}{Fusion Tuning}  & Stage 1-Audio: $5.17\times10^{9}$ & 40 &\multirow{3}{*}{$1.84\times10^{12}$}\\
&&Stage 1-Visual: $1.85\times10^{10}$ & 50\\
&&Stage 2: $2.36\times10^{10}$& 30 \\
\midrule
\multirow{5}{*}{UCF-101}& \multirow{2}{*}{Joint Training} & \multirow{2}{*}{$1.10\times10^{10}$} & \multirow{2}{*}{50} & \multirow{2}{*}{$5.48\times10^{11}$} \\
&&&&\\
\cmidrule{2-5}
&\multirow{3}{*}{Fusion Tuning}  & Stage 1-RGB: $5.48\times10^{9}$ & 20 &\multirow{3}{*}{$7.67\times10^{11}$}\\
&&Stage 1-Flow: $5.48\times10^{9}$ & 80\\
&&Stage 2: $1.10\times10^{10}$& 20 \\
 \bottomrule
\end{tabular}
\caption{Comparison of FLOPs on different datasets.}
\label{tab:flops}
\end{table*}

\subsection{Visualization of dead channels}
\begin{figure*}[t]
  \centering
    \includegraphics[width=14.3cm]{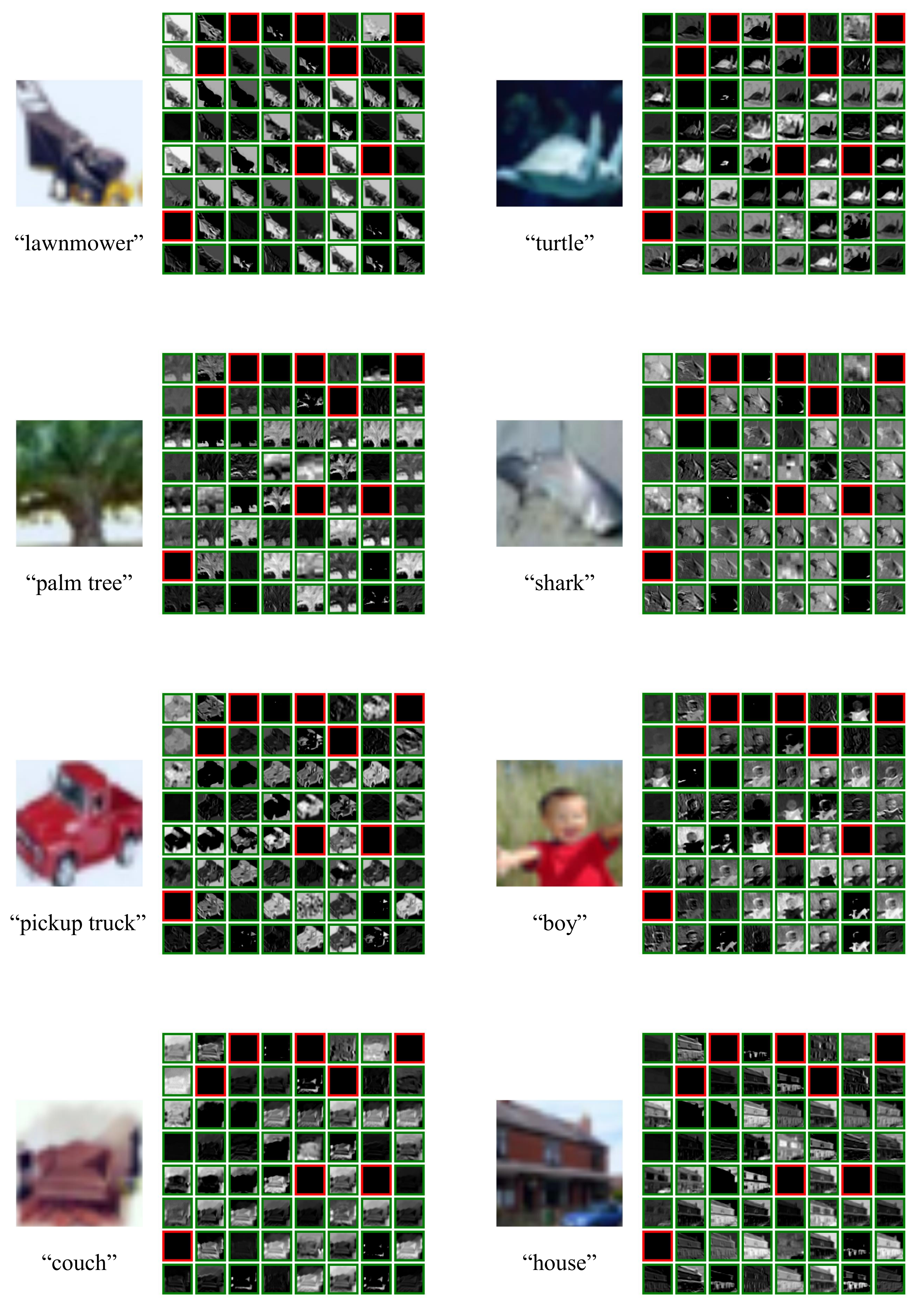}
    \caption{\textbf{Visualization of dead channels in ImageNet pre-trained Resnet-18 model.} The input images are randomly selected from CIFAR-100 dataset. For each pair of images, the input is on the left while the feature maps are on the right. The feature maps of the channels marked with red boxes in the figure are dead features and the corresponding Batchnorm parameters are smaller than $2\times10^{-5}$.}
  \label{fig:dead-sup-img}
\end{figure*}

\begin{figure*}[t]
  \centering
    \includegraphics[width=14.3cm]{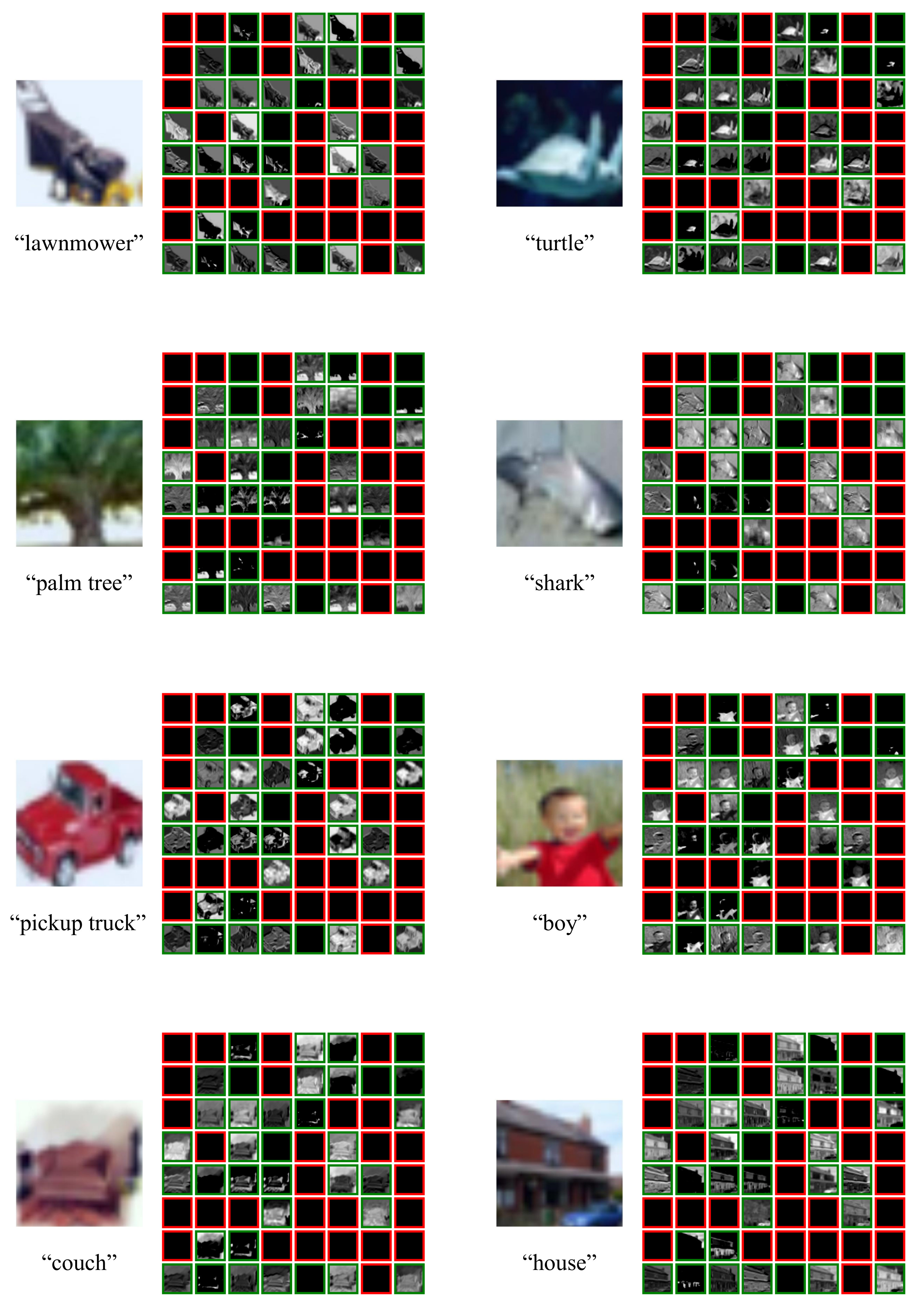}
    \caption{\textbf{Visualization of dead channels in VGGSound pre-trained Resnet-18 model.} The input images are randomly selected from CIFAR-100 dataset. For each pair of images, the input is on the left while the feature maps are on the right. The feature maps of the channels marked with red boxes in the figure are dead features and the corresponding Batchnorm parameters are smaller than $2\times10^{-5}$.}
  \label{fig:dead-sup-vgg}
\end{figure*}

In this part, we show more examples of the dead channels VGGSound and Imagenet pre-trained Resnet-18\cite{he2016deep} models. We randomly select eight images of different classes from CIFAR-100 and show the corresponding feature maps after the first Conv-BN-ReLU. 
Considering the layout, the visualization results \cref{fig:dead-sup-img} and \cref{fig:dead-sup-vgg} are in page 17 and 18.
All the feature maps produced by the same model always have dead features in the same channels, indicating that these channels are hard to be activated. 

\section{Training costs of Fusion Tuning}
Although our proposed Fusion Tuning strategy divides the original joint training into two stages, it does not double the training costs. We calculate the total FLOPs of joint training and our Fusion Tuning strategy on Kinetics-Sounds, AVE (PSP framework~\cite{zhou2021positive}) and UCF-101. As shown in \cref{tab:flops}, the total FLOPs of Fusion Tuning strategy are $1.36\times$, $1.56\times$ and $1.40\times$ total FLOPs of joint training on Kinetic-Sounds, AVE and UCF-101 dataset. This is because the fine-tuned uni-modal encoders in stage one could provide effective initialization for the stage two, so that the fusion step would not use as many epoches as the joint learning strategy but bring better performance. 

\begin{figure*}[t]
  \centering
  \begin{subfigure}{0.49\linewidth}
    \includegraphics[width=8.3cm]{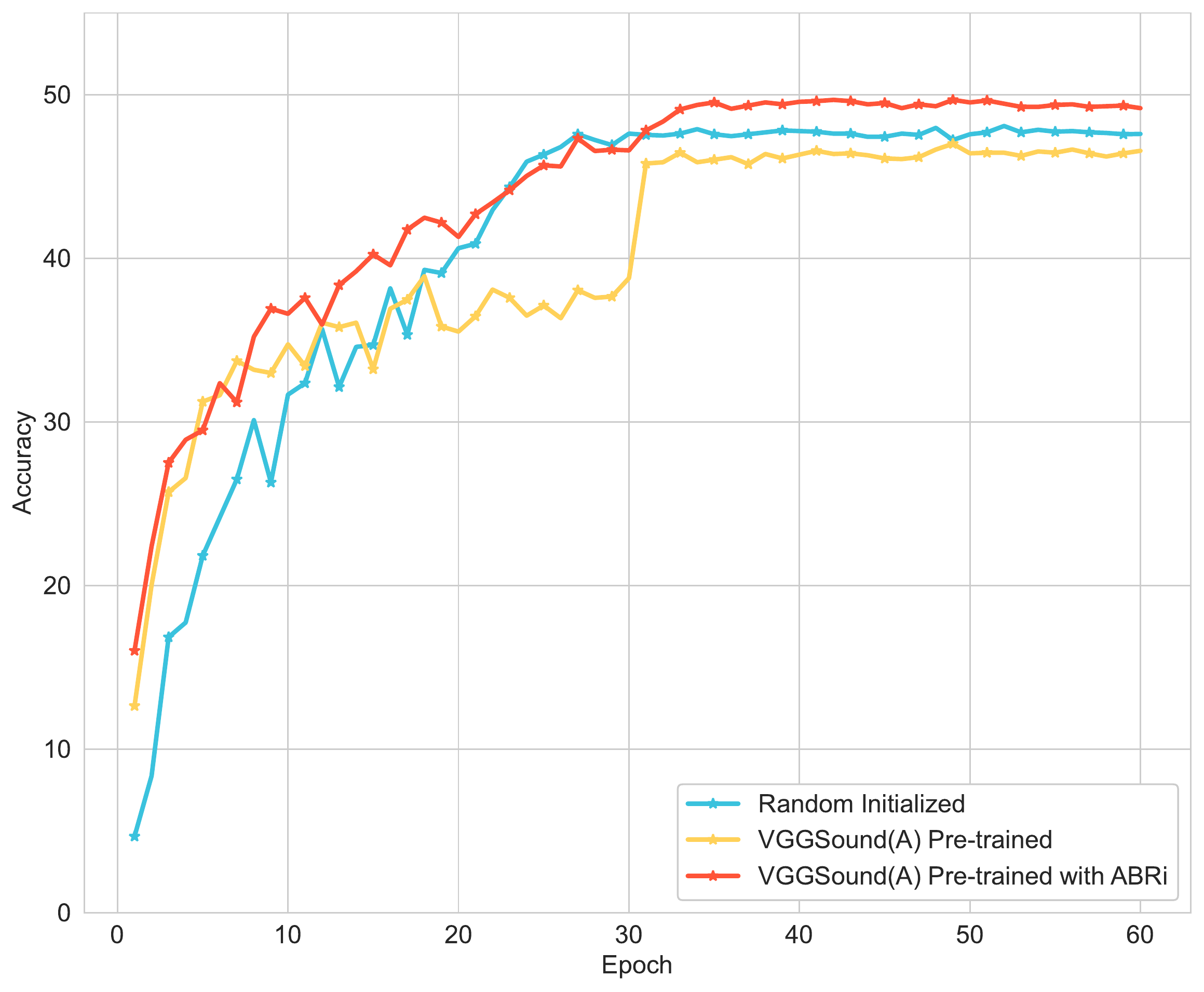}
    \caption{Accuracy of the models on Caltech-256.}
    \label{fig:caltech}
  \end{subfigure}
  \begin{subfigure}{0.49\linewidth}
  \includegraphics[width=8.3cm]{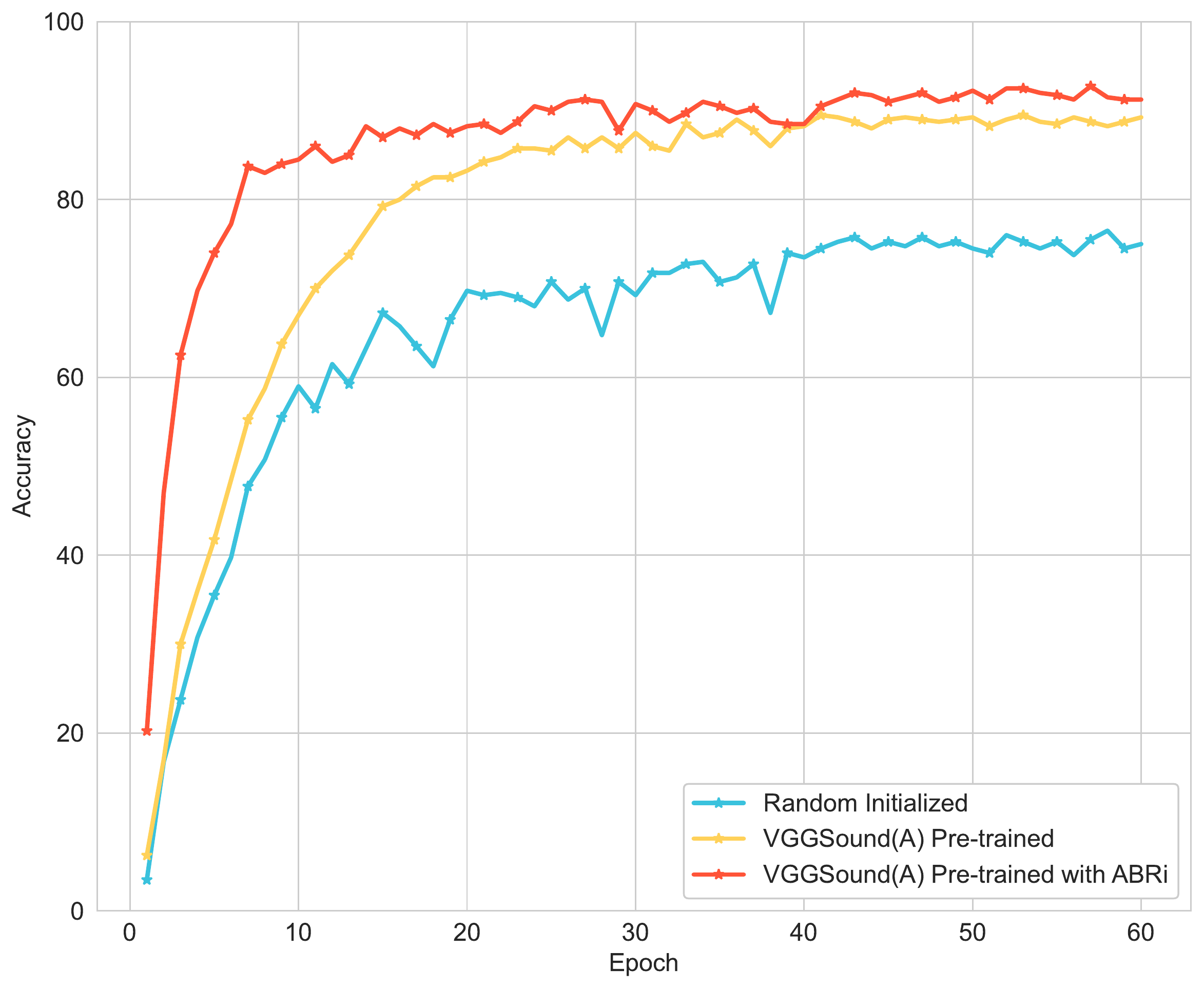}
    \caption{Accuracy of the models on ESC-50 fold-5.}
    \label{fig:esc50}
  \end{subfigure}
  \caption{\textbf{Effect of ABRi on Caltech-256 and ESC-50 dataset.} (a) Performance of random initialized model, VGGSound pre-trained model and VGGSound pre-trained model with ABRi on Caltech-256 testing set. (b) Performance of random initialized model, VGGSound pre-trained model and VGGSound pre-trained model with ABRi on ESC-50 fold-5.}                            
  \label{fig:one2}
\end{figure*}

\section{Supplementary experiment and analysis}
\subsection{Effect of ABRi on uni-modal datasets}
In this part, we show the performance of the random initialized model, VGGSound pre-trained model and VGGSound pre-trained model with ABRi on Caltech-256 and ESC-50 dataset in \cref{fig:one2}. The consistent performance improvement indicates the effectiveness of ABRi on different uni-modal datasets.

\begin{figure}[h]
  \centering
\includegraphics[width=8.4cm]{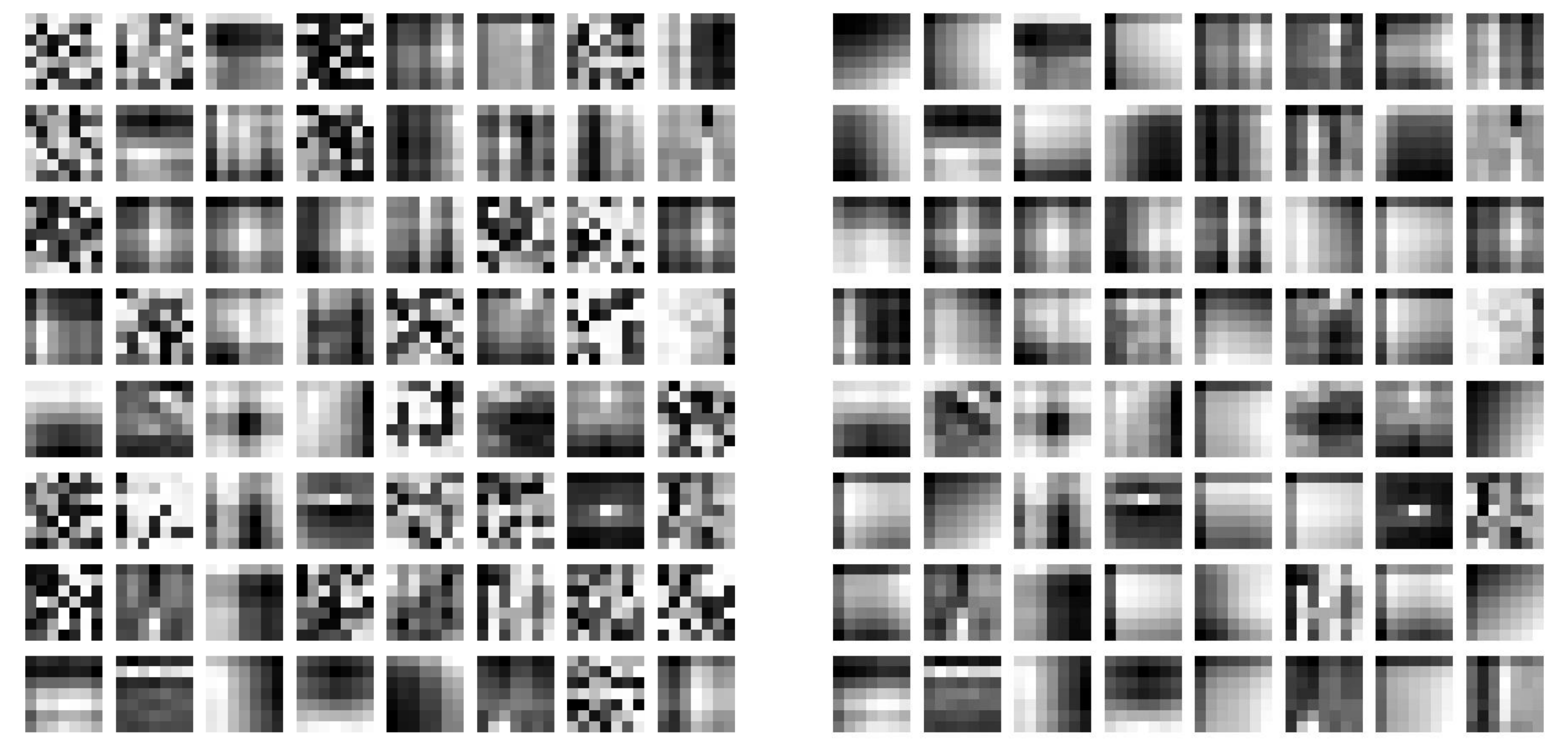}
 \caption{\textbf{Visualization of the filters in the first convolutional layer of VGGSound pre-trained models with and without ABRi.} The models are fine-tuned on CIFAR-100. Left: the model without ABRi. Right: the model with ABRi.}           
 \label{fig:filter}
\end{figure}

We also visualize the filters of the first convolutional layer in VGGSound pre-trained model with and without ABRi after fine-tuning on CIFAR-100. 
As shown in \cref{fig:filter}, many filters in the model without ABRi are still under-optimized after fine-tuning, while most of the filters in the model with ABRi are better updated. This indicates that our ABRi method helps the model to better utilize its capacity.

\subsection{Effect of ABRi on ResNet-34}
We further apply ABRi on ImageNet and VGGSound pre-trained ResNet-34~\cite{he2016deep} models, then evaluate the performance on CIFAR-100 and ESC-50 dataset. The performance improvement shown in \cref{tab:res34 unimodal} proves that ABRi could be extended to more models other than ResNet-18.

\begin{table}[h]
\centering
\tabcolsep=0.25cm
\begin{tabular}{c|c|c}
\toprule
\multirow{2}{*}{Pre-train} &Audio task&Visual task\\
 & ESC50  & CIFAR-100  \\
 \midrule[0.7pt]
Random & 78.55&77.26\\
ImageNet (V) &89.90 &81.67 \\
VGGSound (A)& 92.55 &75.16\\

\midrule
ImageNet (V)\dag & 90.15 (+0.25) &\textbf{81.89 (+0.22)} \\
VGGSound (A)\dag & \textbf{93.80 (+1.25)} & 77.58 (+2.42)\\
 \bottomrule
\end{tabular}
\caption{Comparison of performance with and without ABRi on different uni-modal datasets. The backbone is ResNet-34. \dag \ indicates ABRi is applied.}
\label{tab:res34 unimodal}
\end{table}

\subsection{Effect of Fusion Tuning on uni-modal encoders}
\begin{figure}[h]
  \centering
\includegraphics[width=8.3cm]{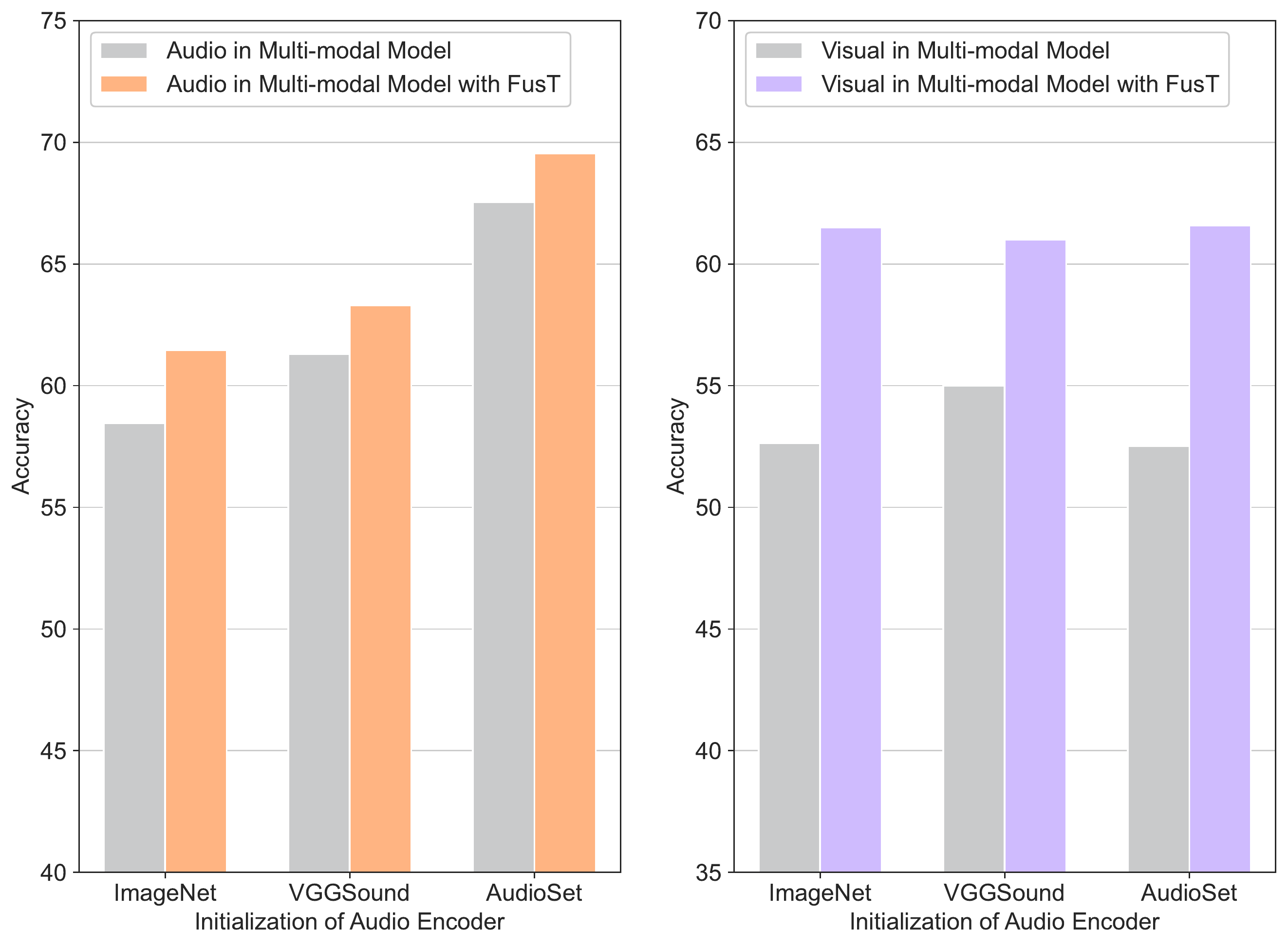}
 \caption{\textbf{Effect of Fusion Tuning on uni-modal encoders.} The experiment is conducted on Kinetics-Sounds dataset. Visual encoders are pre-trained on ImageNet. The accuracy of the uni-modal encoders is obtained through linear-probing.}           
 \label{fig:fust-uni}
\end{figure}

In this part, we show that Fusion Tuning could improve the representation quality of uni-modal encoders and better exploit model knowledge. We compare the accuracy of the multi-modal model and the linear-probing\cite{alain2016understanding} result of the uni-modal in the multi-modal model on joint training and Fusion Tuning. The result shown in \cref{fig:fust-uni} proves the effectiveness of Fusion Tuning. 

\subsection{More results on VGGSound subset}
We compare the Fusion Tuning strategy with the three variants (FusT*, FusT*-lr and FusT-mean) and the two modulation strategies (OGM-GE~\cite{peng2022balanced} and Gradient-Blending\cite{wang2020makes}) on VGGSound subset. The result in \cref{tab:vggbaseline} shows superior performance of our sample-wise method.

\begin{table}[t]
\centering
\tabcolsep=0.26cm
\begin{tabular}{cccc}
\toprule
Framework&Strategy & Pre-train (Audio)&  Acc \\
\midrule[0.7pt]
\multirow{4}{*}{AGVA~\cite{tian2018audio}}&\multirow{2}{*}{FusT\dag} & VGGSound &  71.14 \\
&  & ImageNet & 67.57    \\
\cmidrule(l){2-4}
 &\multirow{2}{*}{FusT\ddag} & VGGSound & \textbf{72.64}  \\
& & ImageNet & \textbf{68.41} \\
\midrule[0.6pt]
\multirow{4}{*}{PSP~\cite{zhou2021positive}}&\multirow{2}{*}{FusT\dag} & VGGSound & 73.43 \\
& & ImageNet  &  69.15 \\
\cmidrule(l){2-4}
 &\multirow{2}{*}{FusT\ddag} & VGGSound & \textbf{74.13} \\
& & ImageNet & \textbf{71.89} \\
 \bottomrule
 
\end{tabular}
\caption{Comparison on AVE dataset with and without Fusion Tuning strategy and ABRi methods in two frameworks. Visual encoders are pre-trained on ImageNet. \dag \ indicates that ABRi is applied on visual encoders. \ddag \ indicates that ABRi is applied on both encoders. }
\label{tab:AVE2}
\end{table}

\begin{table}[t]
\centering
\tabcolsep=0.42cm
\begin{tabular}{ccc}
\toprule
Strategy  & Accuracy& mAP \\
\midrule[0.7pt]
Fusion Tuning\dag & \textbf{84.37} & \textbf{89.28} \\
\midrule
Fusion Tuning\ddag  & 83.62 & 89.00\\
 \bottomrule
 
\end{tabular}
\caption{Performance of the model with Fusion Tuning strategy and ABRi on UCF-101. Encoders are pre-trained on ImageNet.  \dag \ indicates that ABRi is applied on optical flow encoders. \ddag \ indicates that ABRi is applied on both encoders.}
\label{tab:UCF1012}
\end{table}

\begin{table}[t]
\centering
\begin{tabular}{cccccc}
\toprule
$\rho^a$  & $\rho^v$& $\eta^a$& $\eta^v$&$t$&Accuracy \\
\midrule[0.7pt]
1.0 & 0.4 & 1.0 & 1.0 & 0.2 & 77.59 \\
\bottomrule
\end{tabular}
\caption{The anchor setting of hyper-parameters of Fusion Tuning.}
\label{tab:anchor}
\end{table}

\begin{table*}[t]
\centering
\tabcolsep=0.29cm
\begin{tabular}{c|c|ccc|cc|c}
\toprule
&Pre-train (Audio)&FusT* &FusT*-lr  & FusT-mean & OGM-GE\cite{peng2022balanced} & G-Blending\cite{wang2020makes} & FusT \\
 \midrule[0.75pt]
\multirow{2}{*}{Acc}&AudioSet &52.66&52.72&52.71&51.28&52.76&\textbf{52.78}\\
&ImageNet & 49.46&49.53&49.31&47.98&49.56&\textbf{49.62}\\
 \midrule
\multirow{2}{*}{mAP}&
AudioSet &55.21&55.42&55.16&54.61&55.25&\textbf{55.89}\\
&ImageNet &51.13&51.27&51.18&50.42&51.89&\textbf{52.04}  \\
 \bottomrule
 
\end{tabular}
\caption{Comparison with variants of FusT and two modulation strategies on VGGSound subset. Visual encoders are pre-trained on ImageNet. FusT*, FusT*-lr and FusT-mean are three variants of FusT strategy. * indicates masking strategy is not applied in the second stage.}
\label{tab:vggbaseline}
\end{table*}

\begin{figure*}[ht]
  \centering
  \begin{subfigure}{0.49\linewidth}
    \includegraphics[width=8.3cm]{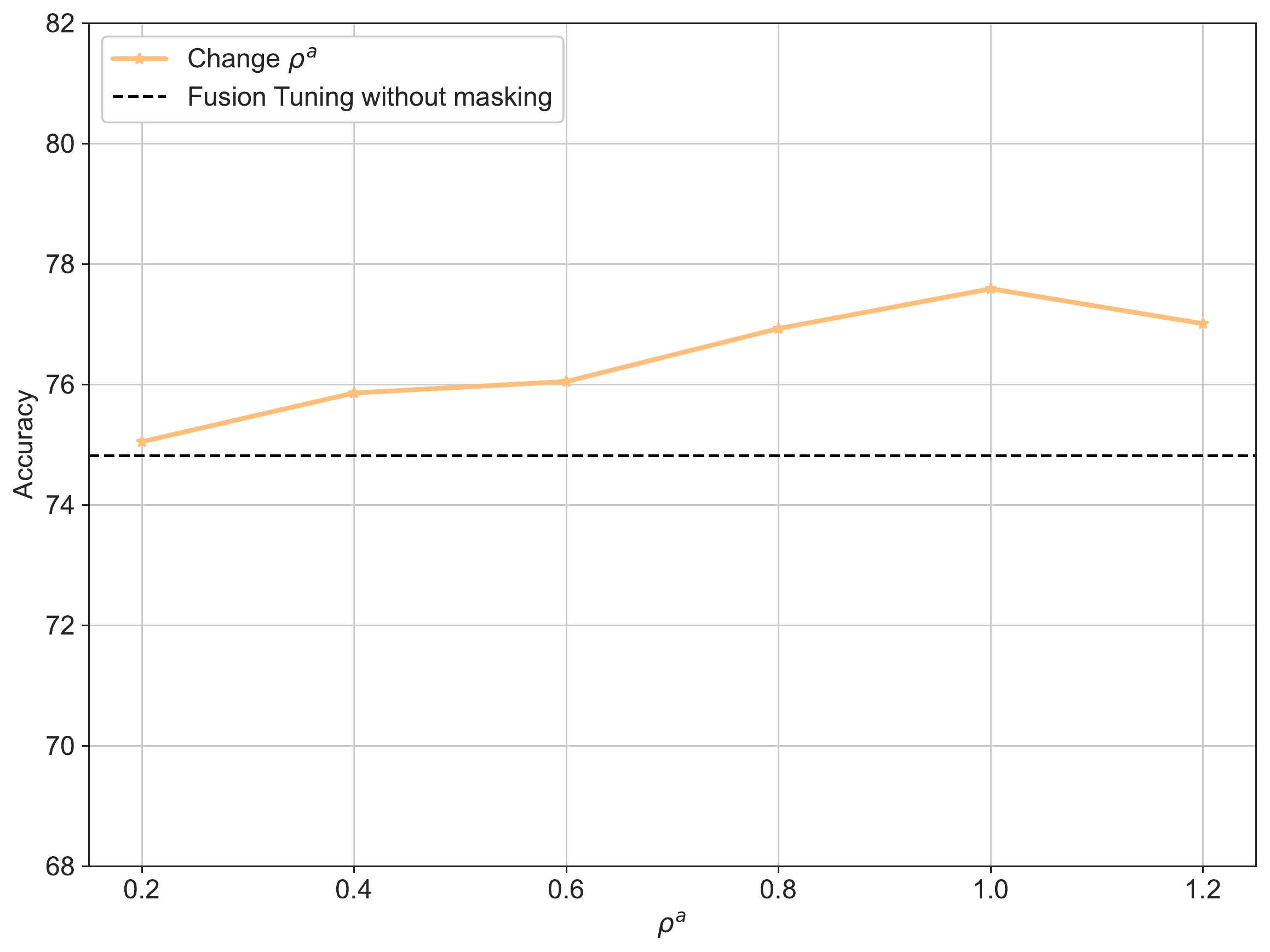}
    \caption{Accuracy of the model when changing $\rho^a$.}
    \label{fig:rhoa}
  \end{subfigure}
  \begin{subfigure}{0.49\linewidth}
  \includegraphics[width=8.3cm]{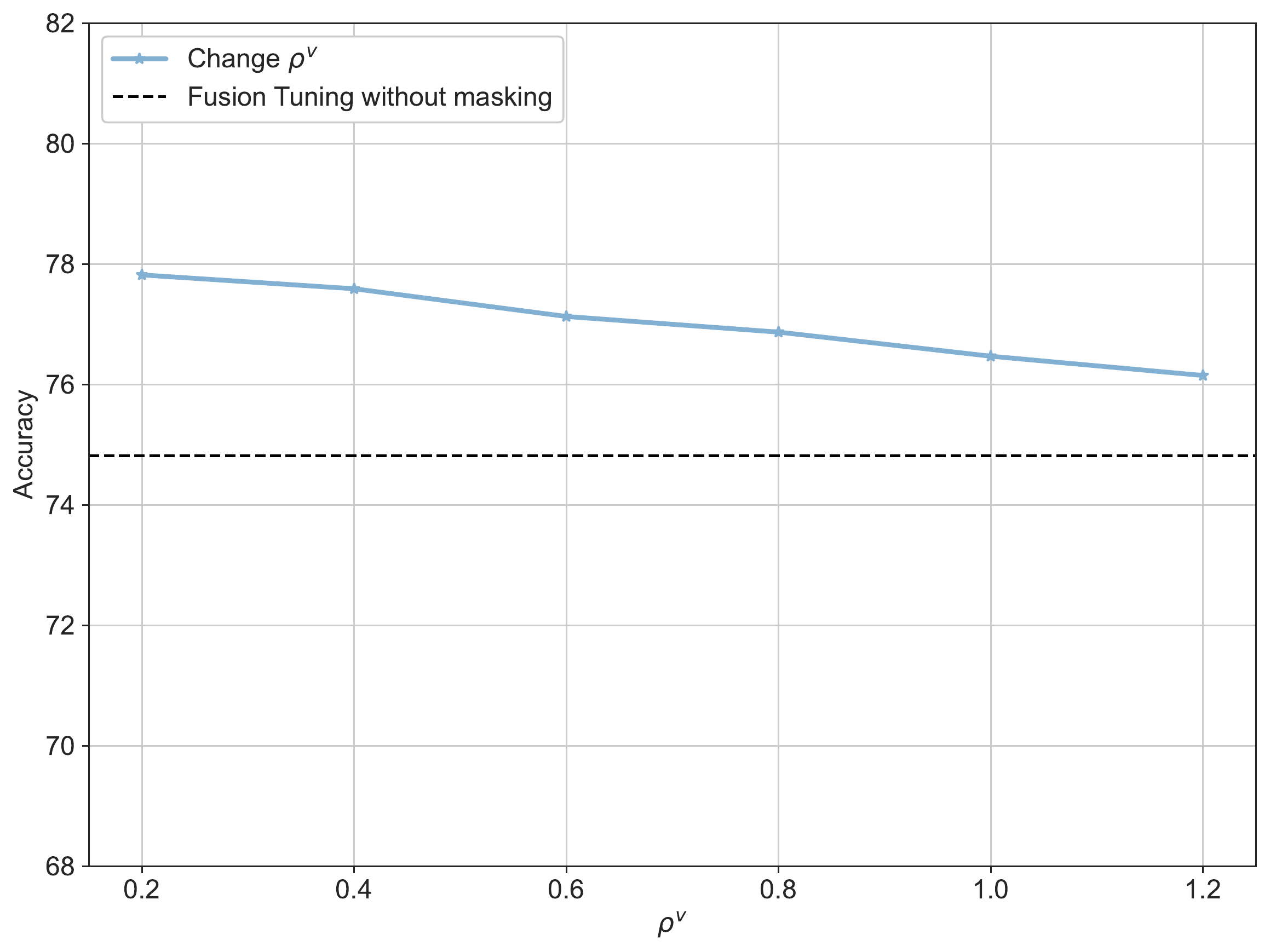}
    \caption{Accuracy of the model when changing $\rho^v$.}
    \label{fig:rhov}
  \end{subfigure}
  \begin{subfigure}{0.49\linewidth}
    \includegraphics[width=8.3cm]{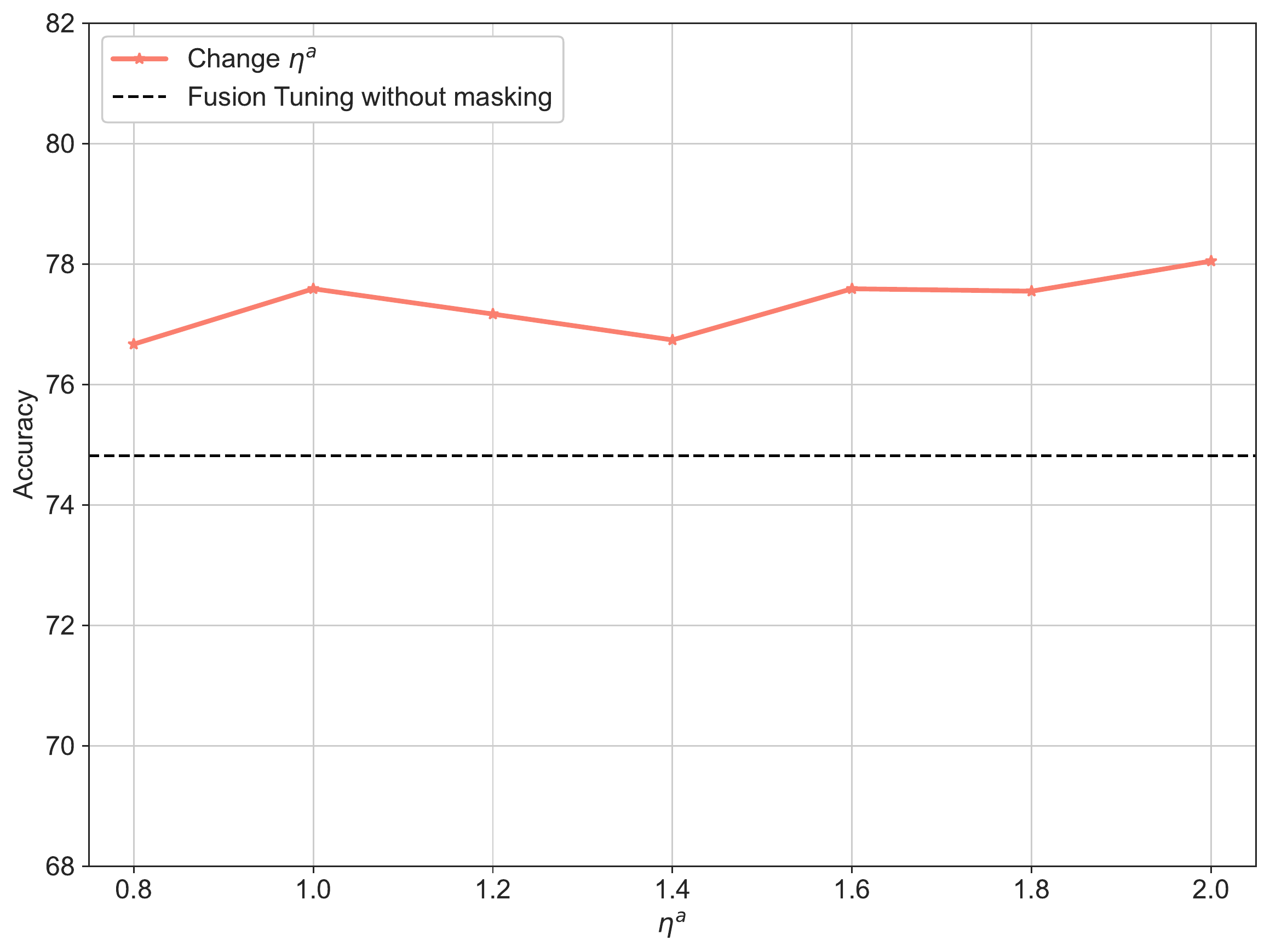}
    \caption{Accuracy of the model when changing $\eta^a$.}
    \label{fig:etaa}
  \end{subfigure}
  \begin{subfigure}{0.49\linewidth}
  \includegraphics[width=8.3cm]{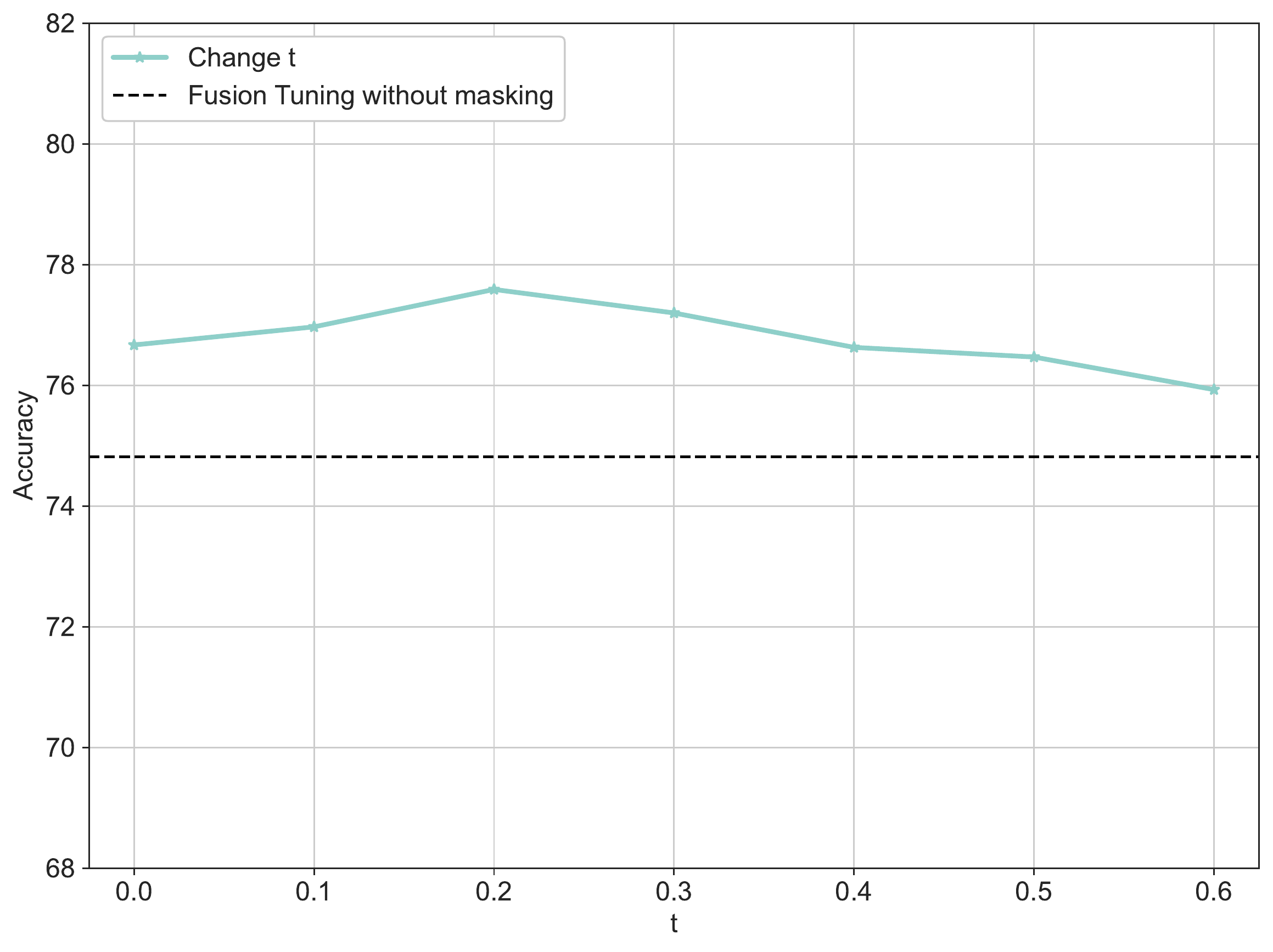}
    \caption{Accuracy of the model when changing $t$.}
    \label{fig:t}
  \end{subfigure}
  \caption{\textbf{Accuracy of the model with different hyper-parameter settings on Kinetics-Sounds.} We evaluate the accuracy when changing $\rho^a$, $\rho^v$, $\eta^a$ and $t$. One parameter changes while the others remain the anchor value in each experiment. The dashed black line marks the accuracy of Fusion Tuning without masking.}                            
  \label{fig:para}
\end{figure*}

\subsection{Integration of the two methods}
We make a comparison of applying ABRi for only one encoder or both encoders on AVE and UCF-101 dataset, as shown in \cref{tab:AVE2} and \cref{tab:UCF1012}. The performance on UCF-101 drops when ABRi is applied on both encoders compared with only on optical flow encoder, while the performance on AVE shows an opposite trend. 
We find that this interesting phenomenon might be associated with the imbalance of modalities in the dataset, which could be reflected by the uni-modal performance in the first stage to some extent. 
When applying ImageNet pre-trained model for both modalities, the accuracy of the two modalities in the first stage has a gap on UCF-101 (77.52 for RGB and 66.02 for optical flow). However, this gap is not evident on AVE dataset (59.46 for audio and 58.72 for visual). 
This indicates that it might be more appropriate to apply ABRi only for the encoder of weaker modality to narrow the imbalance when the gap is evident, while for both encoders when the dataset is more balanced. 
This strategy is similar to the existing modulation methods which aim to bridge the gap in multi-modal joint training~\cite{peng2022balanced,wu2022characterizing}.


\subsection{Evaluation of hyper-parameter settings}

In this part, we test the sensitivity of our Fusion Tuning strategy to different hyper-parameter settings. We choose the setting of the model with ImageNet pre-trained encoders on Kinetics-Sounds dataset shown in \cref{tab:anchor} as the anchor. Then, we evaluate the performance of the model when changing one of the hyper-parameters. The results in \cref{fig:para} indicate that our masking strategy consistently improves the performance under all of these settings. This demonstrates that our Fusion Tuning strategy is not quite sensitive to the hyper-parameter settings.




\section{Introduction of datasets}
\textbf{Pre-training datasets}. 
We select three widely used large-scale datasets: ImageNet-1K~\cite{deng2009imagenet} (visual), VGGSound~\cite{chen2020vggsound} (audio) and AudioSet~\cite{gemmeke2017audio} (audio) to pre-train our models. Imagenet-1K is an image classification dataset with about 1.28M training images and 50k validation images for 1,000 classes. VGGSound is a video dataset containing 309 classes of audio events in everyday life. We successfully download $91\%$ videos of the full dataset. In our pre-training setting, we extract the audio from 168,618 videos in the training set and 13,954 videos in the testing set. AudioSet is a large-scale video dataset with 602 audio event categories. We use the audio clips of 1,990,695 videos in the training set to pre-train the model and evaluate on 20,285 audio clips in the testing set.

\textbf{Uni-modal datasets}. For audio modality, we use ESC-50~\cite{piczak2015esc} and TUT Acoustic Scenes 2016~\cite{mesaros2016tut} datasets. ESC-50 is an audio environment sound classification dataset containing 2,000 audio clips for 50 classes. All the samples are split into five folds and the experiment results are obtained by cross-validation on the five folds. TUT Acoustic Scenes 2016 dataset consists of 6,300 audio clips from 15 acoustic scenes. We train the model on the development set (4,680 samples) and test on the evaluation set (1,620 samples).
We also extract the audio modality from Kinetics-Sounds~\cite{arandjelovic2017look} as a uni-modal dataset.
For visual modality, CIFAR-100~\cite{krizhevsky2009learning} and Caltech-256~\cite{griffin2007caltech} are selected to be the target datasets. CIFAR-100 is a tiny nature image dataset with 100 classes. There is a total of 50,000 images in the training set and 10,000 images in the testing set. Caltech-256 consists of 30,608 images from 256 object categories. The amount of images per category is between 80 and 827.

\textbf{Multi-modal datasets}. For multi-modal tasks, Kinetics-Sounds, AVE~\cite{tian2018audio} and UCF-101~\cite{soomro2012ucf101} are chosen.
Kinetics-Sounds (KS) is a dataset consisting of 31 kinds of human action videos. It contains 15k training samples, 1.9k validation samples and 1.9k testing samples. AVE is an audio-visual event localization dataset with 28 event classes. It contains 4,143 videos of 10 seconds with frame-level annotations.
UCF-101 is an action recognition dataset containing 101 categories of action video. It contains RGB and optical flow modality. The total 13,320 videos are divided into a training set with 9,537 samples and a testing set with 3,783 samples.
We also randomly select a subset of VGGSound, where the amount of samples per class is 90,10,30 in the training, validation and testing set. 

\end{document}